\documentclass{fairmeta}

\usepackage{amsmath,amssymb,amsfonts,amsthm}
\usepackage{adjustbox}
\usepackage{array}
\usepackage{enumitem}
\usepackage{url}

\crefname{appendix}{Appendix}{Appendices}
\Crefname{appendix}{Appendix}{Appendices}

\newtheorem{proposition}{Proposition}
\newcommand{\tcda}{TCDA}
\newcommand{\E}{\mathbb{E}}
\newcommand{\full}{\pi_{\mathrm{full}}}
\newcommand{\Fe}{\mathcal{F}(\varepsilon)}
\newcommand{\pol}[1]{\texttt{#1}}

\title{One Schedule Does Not Fit All: Adaptive Inference Schedules for Diffusion Language Models}

\author[1]{Rabimba Karanjai}
\author[1]{Linsey Pang}
\author[1]{Hemanth Hegadehalli Madhavarao}
\author[1]{Richard Williamson}
\author[1]{Uma Kona}
\author[1]{Prakhar Mehrotra}

\affiliation[1]{PayPal Inc. (\href{https://paypal.com}{paypal.com})}

\contribution{Correspondence: \href{mailto:rkaranjai@paypal.com}{\texttt{rkaranjai@paypal.com}}}
\date{November 2026}

\abstract{
Diffusion language models usually run one fixed denoising schedule for every workload. We show that the cheapest schedule that preserves quality depends on the task, the model and the sampler, and we propose \emph{Task-Calibrated Denoising Allocation} (TCDA), a deployment rule that finds it with an explicit error guarantee. Given a library of candidate samplers (early prefix decoding, state-reusing continuation, reduced time grids, and adaptive samplers such as confidence-threshold decoding), a paired calibration sample, a utility tolerance $\varepsilon$ and degeneration guardrails, TCDA deploys the cheapest candidate whose utility and guardrails are jointly certified by simultaneous one-sided bootstrap bounds, and the full schedule otherwise; the probability of deploying an infeasible schedule is asymptotically at most $\alpha$. Across two model families (the continuous latent flows ELF-B/M/L and the masked diffusion model LLaDA-8B) and seven datasets, TCDA certifies $3.4\times$ fewer denoiser evaluations on XSum and $2\times$ on WMT14 for ELF-B, $8.0\times$ on WMT14 and $4.1\times$ on GSM8K for LLaDA, and $1.33\times$ on OpenWebText for ELF-M and ELF-L, where a protocol frozen before generation confirmed every selection on fresh seeds. False certification stays below $2\%$ in every setting across 500 calibration splits (nominal $5\%$). The certified topology changes across tasks and model scales, and trajectory measurements show why: what matters is when decoded tokens commit, not how much latent discretization error remains. We also find that ELF's reference sampler draws its time grid at random, which makes schedule comparisons on unconditional generation irreproducible unless the grid is fixed.
}

\begin{document}
\maketitle

\section{Introduction}
Diffusion and flow language models generate text by evaluating a large denoiser many times along a trajectory from noise to data. Each model ships with one schedule: Embedded Language Flows (ELF; \citealp{elf2026}) integrate a latent flow over 64 logit-normal time steps, and masked diffusion models such as LLaDA \citep{nie2025llada} unmask a fixed number of tokens per step. A large literature designs better samplers: error-optimized time grids \citep{sabour2024align,chen2024adaptive,gupta2026sharpeuler}, higher-order solvers \citep{li2024accelerating,li2025faster}, and adaptive unmasking \citep{wu2025fastdllm,benhamu2025ebsampler,li2025prophet}. A deployment faces a different question: \emph{for this task and this checkpoint, which of the available samplers can I run, and how much quality do I give up?}

Our data say the answer cannot be fixed in advance. On XSum with ELF-B, decoding the predicted clean latent after 19 of 64 steps is slightly \emph{better} than the full schedule (+0.42 ROUGE-L); on WMT14 the same early decode loses 35 chrF, while continuing from step 19 on a coarser grid keeps quality at half the cost. On LLaDA, committing all remaining tokens at the halfway point costs nothing on WMT14 but 52 accuracy points on GSM8K. On OpenWebText, the certified schedule changes with model scale (\cref{sec:prospective}). Since the right schedule depends on task, model and sampler, it should be chosen per deployment, from data, with error control.

We propose \tcda{}, a certify-then-deploy rule. Given a library of candidate samplers, a calibration sample of $n$ examples run with shared noise, a tolerance $\varepsilon$ on task utility, and guardrails against degenerate text, \tcda{} deploys the cheapest candidate whose utility and guardrails are jointly certified by simultaneous one-sided bootstrap bounds, and falls back to the full schedule otherwise. \tcda{} is not a new sampler: it is a certification layer that can include any sampler, including adaptive and error-optimized ones, as a candidate. Our contributions:
\begin{itemize}
\item \textbf{A certification rule with a guarantee} (\cref{sec:tcda}). We define the target as the cheapest schedule that meets the utility tolerance \emph{and} the guardrails, certify both jointly (intersection--union), and prove that the probability of deploying an infeasible schedule is asymptotically at most $\alpha_u+\alpha_g$ (\cref{prop:fwer}). A max-$T$ statistic with fixed denominators keeps the rule usable for large libraries and near-degenerate candidates, where the studentized statistic breaks down.
\item \textbf{Evidence across two model families, three ELF scales and seven datasets} (\cref{sec:results}), with exact counts over 500 calibration splits, a prospective test whose protocol was hashed before generation, distribution shift, a mechanically generated 127-schedule library, and matched-cost comparisons with error-optimized grids, Heun, adaptive early exit, Fast-dLLM, EB-Sampler and Prophet.
\item \textbf{A mechanism} (\cref{sec:mechanism}). At similar latent error, XSum outputs are settled while WMT outputs are not: token commitment, not discretization error, determines which topology is safe. A decoding-margin proposition makes this precise.
\item \textbf{A reproducibility pitfall} (\cref{sec:grids}). ELF's reference sampler draws its time grid at random once per run; on OpenWebText this moves policy effects by up to 32 standard errors. We use deterministic quantile grids throughout.
\end{itemize}

\section{Models, schedule libraries and time grids}\label{sec:setup-models}
\paragraph{Models.} ELF \citep{elf2026} encodes text with a frozen T5 encoder \citep{raffel2020t5}, integrates a learned velocity field in latent space from $t=0$ (noise) to $t=1$ (data), and maps the result to tokens with a time-conditioned decoder; the predicted clean latent $\hat x_1(z_t)$ is available at every step. Conditional tasks use an ODE sampler with classifier-free guidance \citep{ho2022cfg}; OpenWebText (OWT) uses an SDE sampler. LLaDA-8B-Instruct \citep{nie2025llada} is a masked diffusion model that starts from a fully masked answer and unmasks tokens over a fixed number of steps; Fast-dLLM \citep{wu2025fastdllm} unmasks every token whose confidence exceeds $\tau$, EB-Sampler \citep{benhamu2025ebsampler} unmasks while the summed entropy stays below $\gamma$, and Prophet \citep{li2025prophet} commits all remaining tokens once the top-2 logit gap of the answer exceeds a threshold.

\begin{figure}[t]
\centering
\includegraphics[width=\linewidth]{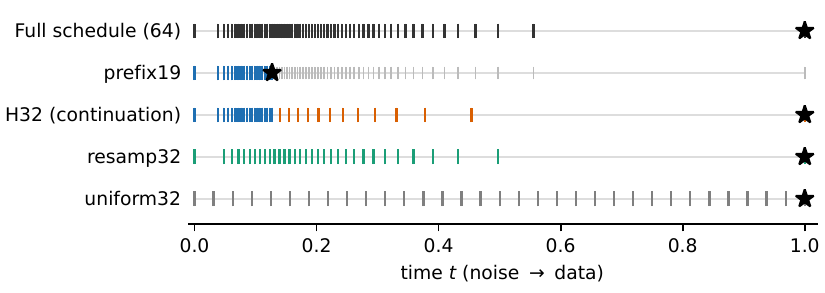}
\vspace{-14pt}
\caption{ELF schedule topologies on the deterministic 64-step grid. Ticks mark denoiser evaluations and $\star$ the decode point. \pol{prefix19} decodes the predicted clean latent after 19 steps (grey: skipped steps); \pol{H32} reuses the first 19 steps and continues on 13 coarser steps.}
\label{fig:topologies}
\vspace{-6pt}
\end{figure}

\paragraph{Schedule topologies.} The cost $c(\pi)$ of a schedule is its number of denoiser evaluations (NFE). For ELF (\cref{fig:topologies}): \pol{prefixK} runs the first $K$ steps of the full schedule and decodes $\hat x_1(z_{t_K})$; continuation \pol{HH} runs the full schedule to step $p$ ($p=19$ for conditional tasks, $p=H/2$ on OWT) and continues to $t=1$ on a coarser grid of $H-p$ steps, reusing the latent and self-conditioning state; \pol{resampK} and \pol{uniformK} traverse $[0,1]$ with $K$ logit-normal or uniform steps. For LLaDA we use early commit (\pol{commitK}: unmask everything after $K$ steps), uniform, front- and back-loaded step allocations, a hybrid, and the three adaptive samplers at several thresholds (21 candidates per task). All candidates share the initial noise of the full schedule, so differences are paired.

\paragraph{Random versus deterministic time grids.}\label{sec:grids}
ELF's reference implementation places its $K$ steps at sorted random draws from the logit-normal training distribution ($P_{\rm mean}=-1.5$, $P_{\rm std}=0.8$), drawn once per run. A calibration run therefore certifies one random grid, and a deployment with another seed uses a different one. To measure the effect we held the initial noise fixed and redrew only the grid (8 draws, 256 examples): the spread of a policy's effect across grid draws is 0.5--1.4 within-run standard errors on XSum and WMT, but 14--32 on ELF-B OWT and 1.4--17 on ELF-M/L OWT, and one of the eight draws makes ELF-B's full OWT schedule degenerate (Gen.\ PPL 7.8, unigram entropy 2.9; \cref{app:grids}). We therefore use deterministic quantile grids, $t_k=\sigma\!\big(P_{\rm mean}+P_{\rm std}\,\Phi^{-1}(k/K)\big)$, for every topology. They reproduce the random-grid quality of the full schedule for ELF-B XSum and WMT (28.17 vs 28.10 ROUGE-L; 52.59 vs 52.65 chrF) and for ELF-M/L OWT (Gen.\ PPL 21.5 and 23.3 vs 20.1--23.2 and 21.5--24.7 across random draws), but not for ELF-B OWT (51.3 vs 21--32). We therefore make no speed claim for ELF-B OWT.

\section{Task-Calibrated Denoising Allocation}\label{sec:tcda}
\paragraph{Target.} Let $u_i(\pi)$ be the utility of schedule $\pi$ on example $i$ (ROUGE-L$\times100$, chrF, exact match, or $-$NLL under GPT-2 Large) and $\Delta_i(\pi)=u_i(\pi)-u_i(\full)$. Degeneration is measured by guardrail contrasts $h_{ij}(\pi)$ with thresholds $\tau_j$, e.g.\ the change in output length relative to $\full$, empty outputs, adjacent repetition, distinct-2 and decoder entropy (\cref{app:guardrails}). The feasible set and the target are
\begin{equation}
\Fe=\{\full\}\cup\big\{\pi:\ \E\Delta(\pi)\ge-\varepsilon\ \text{and}\ \E h_j(\pi)\le\tau_j\ \forall j\big\},\qquad \pi^*(\varepsilon)=\arg\min_{\pi\in\Fe}c(\pi).
\end{equation}
Defining feasibility with the guardrails matters: on OWT, schedules that loop or repeat lower the NLL of GPT-2 and look \emph{better} by utility alone.

\paragraph{Certification rule.} On a calibration sample of $n$ examples, let $\hat\mu_\pi$ and $\hat s_\pi=\mathrm{sd}(\Delta(\pi))/\sqrt n$ be the mean and standard error of $\Delta(\pi)$, and $\hat\eta_{\pi j},\hat s_{\pi j}$ those of $h_j(\pi)$. Draw $B$ bootstrap resamples of the calibration indices and compute
\begin{equation}
c_u=Q_{1-\alpha_u}\Big[\max_{\pi:\hat s_\pi>0}\frac{\hat\mu^{*}_{\pi}-\hat\mu_\pi}{\hat s_\pi}\Big],\qquad
c_g=Q_{1-\alpha_g}\Big[\max_{(\pi,j):\hat s_{\pi j}>0}\frac{\hat\eta^{*}_{\pi j}-\hat\eta_{\pi j}}{\hat s_{\pi j}}\Big],
\end{equation}
where $Q_q$ is the bootstrap $q$-quantile and the denominators are the \emph{original-sample} standard errors. A candidate is certified if $\hat s_\pi>0$, $\mathrm{LCB}_\pi=\hat\mu_\pi-c_u\hat s_\pi\ge-\varepsilon$, and $\mathrm{UCB}_{\pi j}=\hat\eta_{\pi j}+c_g\hat s_{\pi j}\le\tau_j$ for every guardrail (contrasts with $\hat s_{\pi j}=0$ are checked at their point estimate). \tcda{} deploys the cheapest certified candidate (ties: higher $\hat\mu_\pi$) and $\full$ if none is certified. We use $\alpha_u=0.04$, $\alpha_g=0.01$ and $B=2{,}000$.

The submitted implementation certified utility only, studentizing each bootstrap replicate by its own standard error; the guardrails entered the reference policy but not the per-split certification. Both choices fail in practice. Without guardrails in the bound, a looping OWT candidate was certified in 36 of 500 calibration splits ($n=128$) on the submission's data and in all 500 under deterministic grids (\cref{sec:elfb}). And a replicate with zero variance (frequent when two schedules agree on most examples, or for binary metrics) makes the studentized maximum infinite, which forces needless fallback: the submitted rule falls back in 406/500 splits with the dense XSum library and never deploys Fast-dLLM on LLaDA WMT14, where \tcda{} deploys it at $7.5\times$; a studentized joint rule we used in our first prospective run reached critical values up to 306 instead of $\approx2.6$. Fixed denominators keep the critical value bounded with the same asymptotics, and joint certification removes the false certifications.

\begin{proposition}[False certification]\label{prop:fwer}
Let the calibration examples be i.i.d.\ from the deployment distribution, the library be finite, and $\Delta(\pi)$ and $h_j(\pi)$ have finite second moments. Then, for contrasts with positive variance,
$\limsup_{n\to\infty}\Pr\big(\hat\pi\notin\Fe\big)\le\alpha_u+\alpha_g$, where $\hat\pi$ is the deployed schedule.
\end{proposition}
\emph{Proof sketch} (full proof in \cref{app:proofs}). If $\hat\pi\ne\full$ is infeasible, either $\E\Delta(\hat\pi)<-\varepsilon\le\mathrm{LCB}_{\hat\pi}$, which implies $\max_\pi(\hat\mu_\pi-\E\Delta(\pi))/\hat s_\pi>c_u$, or some $\E h_j(\hat\pi)>\tau_j\ge \mathrm{UCB}_{\hat\pi j}$, which implies the analogous event for the guardrails. By the bootstrap central limit theorem the two maxima exceed $c_u$ and $c_g$ with limiting probabilities at most $\alpha_u$ and $\alpha_g$; a union bound finishes the proof. The event does not depend on which certified candidate is deployed, so choosing the cheapest is free. $\square$

The guarantee is relative to the library and to exchangeable calibration and deployment inputs; \cref{sec:shift} tests the latter. \tcda{} relates to Learn-then-Test \citep{angelopoulos2021learn} and CALM \citep{schuster2022calm}, which calibrate thresholds for risk control; it differs in testing paired non-inferiority of expected utility with guardrails and in choosing among sampler topologies rather than a scalar threshold.

\section{Experimental setup}\label{sec:exp}
\paragraph{Tasks and models.} ELF-B: XSum \citep{narayan2018xsum} (1,000 validation examples, ROUGE-L), WMT14 De$\to$En \citep{bojar2014wmt} (3,000, sentence chrF; \citealp{popovic2015chrf}) and OWT \citep{gokaslan2019openwebtext} (1,000 unconditional samples, $-$NLL under GPT-2 Large; \citealp{radford2019gpt2}). ELF-M/L: OWT, public checkpoints. ELF-M XSum: a 342M-parameter model we trained (\cref{app:elfm}). LLaDA-8B-Instruct: GSM8K \citep{cobbe2021gsm8k} (1,000 test questions, exact match), XSum and WMT14 (1,000 each). Full schedules use 64 steps (ELF XSum/WMT, ELF-M/L OWT, LLaDA XSum), 32 (ELF-B OWT), 128 (LLaDA WMT14) and 256 (LLaDA GSM8K).

\paragraph{Libraries.} The \emph{core} ELF-B library has 11 candidates per conditional task (prefix decoding at 19--32 steps, continuation \pol{H32/40/48}, resampled and uniform grids at 32--48 steps) and 10 for OWT; the prospective OWT library has 14 per model. The \emph{dense} library is generated mechanically: prefix decoding at every step, resampled and uniform grids every 4 steps, and continuation over 11 budgets $\times$ 3 branch points (127 candidates for XSum and WMT, including the core continuation schedules).

\paragraph{Protocol.} For each task we draw 500 random calibration sets of size $n\in\{64,128,256\}$ for ELF and $n\in\{128,256\}$ for LLaDA ($n=512$ for GSM8K) from the pool and run \tcda{} on each. We report how often the deployed schedule equals $\pi^*(\varepsilon)$ computed on the full pool, the mean deployed cost, \emph{false certifications} (deployed $\notin\Fe$ on the full pool) with Clopper--Pearson intervals, and held-out violations. Wall-clock times use preloaded weights, a warm-up run, 7 interleaved repetitions and explicit CUDA synchronization, excluding metric computation.

\section{Results}\label{sec:results}
\subsection{ELF-B: the certified schedule depends on the task}\label{sec:elfb}
\Cref{tab:elfb} shows the deployed schedules at $n=128$ and \cref{tab:reliability} the full tolerance range. On XSum, \tcda{} deploys \pol{prefix19} in 499/500 splits ($3.37\times$ fewer evaluations, $3.28\times$ wall-clock); on WMT, \pol{H32} in 466/500 ($2.0\times$). False certification is 0/500 in both cells and at most 5/500 over all 30 XSum and WMT settings ($n\in\{64,128,256\}$, $\varepsilon\in\{0,0.25,0.5,1,2\}$; \cref{app:fulltables}). Under tight tolerances \tcda{} falls back, as intended (WMT $\varepsilon=0$: 495/500 splits). The same holds on the original random-grid data of the submission (XSum $\varepsilon=1$: \pol{prefix19} 500/500; WMT: \pol{H32} 486/500; at most 7/500 false certifications in any setting; \cref{app:fulltables}).

\emph{Guardrails are necessary.} On ELF-B OWT, a utility-only rule certifies \pol{hybrid\_H16}, whose text loops (distinct-2 0.10 vs 0.75), in 500/500 splits; the joint rule never does (0/500 in all 12 OWT settings).

\emph{Larger libraries help.} With the dense library, the certified speedup at $\varepsilon=1$ rises to $6.75\times$ on XSum (modal \pol{prefix9}; \pol{prefix8} already matches the full schedule) and $3.8\times$ on WMT (continuation with 16 evaluations), with at most 3/500 false certifications in any setting. Max-$T$ accounts for the correlation between neighbouring schedules, so a larger library does not erase the gain; the submitted studentized rule, by contrast, falls back in 406/500 XSum splits with this library.

\begin{table}[t]
\centering\small\setlength{\tabcolsep}{4pt}
\caption{ELF-B, deterministic grids, $n=128$, 500 splits. $\Delta$: full-pool mean difference to the full schedule. Selected: splits deploying the modal schedule. False cert.: deployed schedule infeasible on the full pool. Wall-clock on 1,000 examples, RTX PRO 6000; $^\dagger$B200, measured for the submission.}
\label{tab:elfb}
\begin{adjustbox}{max width=\linewidth}
\begin{tabular}{llllrrrr}
\toprule
Task, $\varepsilon$ & Library & Deployed & $\Delta$ (full pool) & Selected & False cert. & NFE ratio & Wall-clock\\
\midrule
XSum, 1 ROUGE-L & core & \pol{prefix19} & $+0.42$ & 499/500 & 0/500 & $3.37\times$ & $3.28\times$\\
                & dense & \pol{prefix9} (modal) & $+0.28$ & 130/500 & 0/500 & $6.75\times$ & --\\
WMT14, 1 chrF   & core & \pol{H32} & $+0.06$ & 466/500 & 0/500 & $2.00\times$ & $1.94\times^\dagger$\\
                & dense & \pol{H16} (branch 12) & $+0.22$ & 275/500 & 1/500 & $3.81\times$ & --\\
\bottomrule
\end{tabular}
\end{adjustbox}
\vspace{-6pt}
\end{table}

\begin{table}[t]
\centering\small
\caption{Reliability at $n=128$ over 500 calibration splits (ELF-B, deterministic grids, core library). Deployed: modal schedule and its count; false: splits whose deployed schedule is infeasible on the full pool. The submitted rule certifies utility only; on OWT it deploys a looping schedule in every split.}
\label{tab:reliability}
\begin{adjustbox}{max width=\linewidth}
\begin{tabular}{llllrlrr}
\toprule
 & & & \multicolumn{2}{c}{Submitted rule (utility only)} & \multicolumn{3}{c}{\tcda{}} \\ \cmidrule(lr){4-5} \cmidrule(lr){6-8}
Task & $\varepsilon$ & $\pi^*(\varepsilon)$ & deployed (modal) & false & deployed (modal) & false & NFE ratio \\
\midrule
XSum & 0 & \texttt{prefix19} & \texttt{canonical64} 353 & 0 & \texttt{canonical64} 382 & 0 & 1.17$\times$ \\
 & 0.25 & \texttt{prefix19} & \texttt{prefix19} 261 & 0 & \texttt{prefix19} 257 & 0 & 2.36$\times$ \\
 & 0.5 & \texttt{prefix19} & \texttt{prefix19} 426 & 0 & \texttt{prefix19} 440 & 0 & 3.17$\times$ \\
 & 1 & \texttt{prefix19} & \texttt{prefix19} 485 & 0 & \texttt{prefix19} 499 & 0 & 3.37$\times$ \\
 & 2 & \texttt{prefix19} & \texttt{prefix19} 494 & 0 & \texttt{prefix19} 500 & 0 & 3.37$\times$ \\
\midrule
WMT14 & 0 & \texttt{H32} & \texttt{canonical64} 493 & 0 & \texttt{canonical64} 495 & 0 & 1.00$\times$ \\
 & 0.25 & \texttt{H32} & \texttt{H32} 148 & 2 & \texttt{H32} 177 & 2 & 1.55$\times$ \\
 & 0.5 & \texttt{H32} & \texttt{H32} 352 & 0 & \texttt{H32} 428 & 0 & 1.94$\times$ \\
 & 1 & \texttt{H32} & \texttt{H32} 459 & 0 & \texttt{H32} 466 & 0 & 2.00$\times$ \\
 & 2 & \texttt{H32} & \texttt{H32} 464 & 0 & \texttt{H32} 464 & 0 & 2.00$\times$ \\
\midrule
OWT (ELF-B) & 5\% & \texttt{uniform24} & \texttt{hybrid\_H16} 500 & 500 & \texttt{canonical32} 482 & 0 & 1.01$\times$ \\
 & 10\% & \texttt{uniform24} & \texttt{hybrid\_H16} 500 & 500 & \texttt{canonical32} 292 & 0 & 1.12$\times$ \\
 & 15\% & \texttt{uniform24} & \texttt{hybrid\_H16} 500 & 500 & \texttt{uniform24} 424 & 0 & 1.27$\times$ \\
 & 20\% & \texttt{uniform24} & \texttt{hybrid\_H16} 500 & 500 & \texttt{uniform24} 492 & 0 & 1.33$\times$ \\
\bottomrule
\end{tabular}
\end{adjustbox}

\vspace{-6pt}
\end{table}

\subsection{Beyond ELF: LLaDA-8B}\label{sec:llada}
On LLaDA the pattern is different (\cref{tab:llada}). Early commit is near-lossless on WMT14 at $2\times$ ($-0.005$ chrF) but loses 51.9 accuracy points on GSM8K; the cheapest feasible schedule is Fast-dLLM with $\tau=0.8$ on GSM8K and WMT14, and a static back-loaded 16-step schedule on XSum. \tcda{} certifies $8.0\times$ fewer forward passes on WMT14 ($7.5\times$ wall-clock) and $4.1\times$ on GSM8K, with 0/500 false certifications in every reported cell and at most 1/500 elsewhere. With GSM8K's binary utility, a 5-point tolerance needs $n=512$ for Fast-dLLM to be deployed in most splits (398/500, vs.\ 166/500 at $n=256$). A variant that certifies utility only and checks guardrails at their point estimates produced up to 10/500 false certifications on XSum, again showing the value of joint certification.

\begin{table}[t]
\centering\small\setlength{\tabcolsep}{3.6pt}
\caption{LLaDA-8B-Instruct, 21 candidates per task, 500 splits. $\pi^*$: cheapest feasible schedule on the full pool. Fast-dLLM's default ($\tau=0.9$) is in \cref{app:llada}.}
\label{tab:llada}
\begin{adjustbox}{max width=\linewidth}
\begin{tabular}{lllllrrr}
\toprule
Task (full) & $\varepsilon$, $n$ & $\pi^*(\varepsilon)$, $\Delta$ & Deployed & Selected & False cert. & NFE ratio & Wall-clock\\
\midrule
GSM8K (77.9\% EM) & 5 pts, 512 & \pol{fd0.8}, $-0.7$ & \pol{fd0.8} & 398/500 & 0/500 & $4.05\times$ & $4.25\times$\\
XSum (19.09)      & 0.5, 128 & \pol{back16}, $-0.21$ & \pol{fd0.8} & 424/500 & 0/500 & $1.70\times$ & $1.70\times$\\
WMT14 (55.90)     & 1, 256 & \pol{fd0.8}, $-0.21$ & \pol{fd0.8} & 451/500 & 0/500 & $8.03\times$ & $7.53\times$\\
\bottomrule
\end{tabular}
\end{adjustbox}
\vspace{-6pt}
\end{table}

\subsection{Prospective tests on fresh seeds}\label{sec:prospective}
\paragraph{OWT at three scales.} We froze the protocol, library, guardrails and rules, hashed the protocol, and then generated: calibration on noise seeds 500/501 (256 samples), and confirmation of every schedule selected by the primary or secondary rule on seeds 600--602 ($3\times1{,}000$ samples). A selection is confirmed if all three per-seed one-sided 95\% bounds and a hierarchical bound (seeds, then samples) clear $-\varepsilon$ and all guardrails pass. Every selection was confirmed (\cref{tab:prospective}): \pol{prefix48} on ELF-M and continuation \pol{H48} on ELF-L, each $1.33\times$ fewer evaluations and $1.33\times$ wall-clock, already at the 5\% perplexity tolerance. The certified topology differs between scales of the same model family on the same task. A one-way ANOVA across the three seeds gives $p=0.034$ for ELF-M and $0.12$ for ELF-L (none significant after Holm correction over four tests; the ELF-M spread is 0.005 nats, a tenth of the tolerance). Our first prospective run used ELF's random grids; it confirmed three of seven distinct selections (including ELF-M \pol{prefix48}, and ELF-L \pol{H48} at 10--20\% but not 5\%) and failed four, which is how we found the random-grid problem (\cref{app:e1}).

\begin{table}[t]
\centering\small\setlength{\tabcolsep}{4pt}
\caption{Prospective OWT confirmation (deterministic grids; tolerance 5\% relative PPL, $\varepsilon=0.049$ nats). $\Delta$ in nats of $-$NLL (GPT-2 Large) per confirmation seed; hierarchical one-sided 95\% lower bound; ELF Gen.\ PPL and unigram entropy pooled over 3{,}000 samples.}
\label{tab:prospective}
\begin{adjustbox}{max width=\linewidth}
\begin{tabular}{llrrrrllr}
\toprule
Model & Deployed & \multicolumn{3}{c}{$\Delta$, seeds 600 / 601 / 602} & Hier.\ LCB & Gen.\ PPL & Entropy & Wall-clock\\
\midrule
ELF-M & \pol{prefix48} & $-0.006$ & $-0.000$ & $-0.001$ & $-0.005$ & 21.33 vs 21.23 & 5.19 vs 5.19 & $1.33\times$\\
ELF-L & \pol{H48}      & $-0.029$ & $-0.022$ & $-0.029$ & $-0.030$ & 23.68 vs 23.03 & 5.29 vs 5.28 & $1.33\times$\\
\bottomrule
\end{tabular}
\end{adjustbox}
\vspace{-6pt}
\end{table}

\paragraph{XSum on an independently trained ELF-M.} A single preregistered candidate, \pol{prefix19}, was tested on 1,000 untouched XSum examples with three fresh seeds at $\varepsilon=0.5$: hierarchical $\Delta=+0.18$ ROUGE-L (one-sided lower bound $+0.10$), all guardrails pass, $3.26\times$ wall-clock (\cref{app:elfm}). The early-decoding regime transfers across scale and training run. This test originally used random grids; we repeated it with the deterministic grid under a protocol frozen before generation. On 1,000 never-used examples (IDs 2000--2999) with fresh seeds, per-seed $\Delta=+0.39,+0.22,+0.34$ (all lower bounds $\ge+0.09$; hierarchical $+0.32$, bound $+0.22$), all guardrails pass; on the original cohort and seeds the deterministic grid changes per-seed $\Delta$ by at most 0.03. Across 8 random grids with fixed noise, the spread of $\Delta$ is 0.25 within-run standard errors (\cref{app:elfm}).

\subsection{Comparison with other accelerations at matched cost}\label{sec:baselines}
\begin{figure}[t]
\centering
\includegraphics[width=\linewidth]{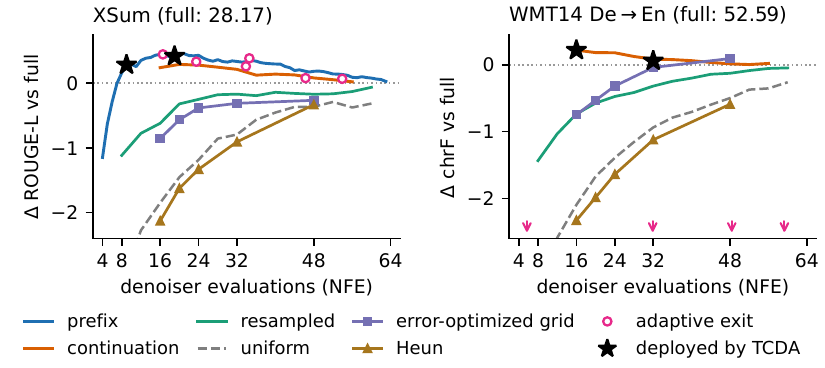}
\vspace{-14pt}
\caption{ELF-B quality vs.\ cost under deterministic grids (full pools). Stars: schedules deployed by \tcda{} at $\varepsilon=1$ (core and dense library). Off scale: WMT prefix decoding ($-27$ to $-49$ chrF), uniform grids at 8 NFE, and WMT adaptive exit (arrows; $-5.7$ to $-45.6$). Not shown: Heun on the logit-normal grid (XSum $-6$ to $-9$; WMT $-10$ to $-19$).}
\label{fig:frontier}
\vspace{-6pt}
\end{figure}

\Cref{fig:frontier} compares the ELF-B library with three alternatives at matched NFE: an error-optimized step grid in the style of SharpEuler \citep{gupta2026sharpeuler} (step density $\propto$ local velocity sharpness$^{1/2}$, profiled on 128 held-out trajectories; the flow analogue of \citealp{chen2024adaptive}), Heun's second-order method, and per-example adaptive exit (stop when decoded tokens are unchanged for $s$ steps, or when the 10th-percentile decoder margin exceeds a profiled threshold). On XSum, prefix decoding dominates at every budget (20 NFE: $+0.46$ ROUGE-L vs $-0.57$ for the error-optimized grid and $-1.6$ for Heun on a uniform grid; $-8.2$ on the logit-normal grid). On WMT, the error-optimized grid is competitive at 32--48 NFE ($-0.04$ vs $+0.06$ for \pol{H32}) but not at 16 ($-0.75$ vs $+0.22$ for continuation). Adaptive exit is good on XSum ($+0.45$ at 16.6 NFE) and unsafe on WMT ($-5.7$ to $-45.6$ chrF): a per-example rule that works on one task fails on another, which is the case for per-task certification. Measured wall-clock on XSum follows NFE for fixed schedules (\pol{prefix19} $3.28\times$, error-optimized 20 steps $3.12\times$, Heun-20 $3.11\times$) but not for adaptive exit, whose per-step decoding costs time (stability exit $1.88\times$ at $2.72\times$ fewer NFE; margin exit $0.85\times$). On LLaDA, Fast-dLLM and EB-Sampler are candidates inside \tcda{}; the official Prophet implementation reproduces our simulation on GSM8K (51.5\% vs 50.5\% at $\approx1.4\times$) and, with its evaluation script's answer-prefix constraint, reaches 73.0\% at $1.22\times$, below Fast-dLLM and EB-Sampler on the same questions (\cref{app:llada}).

\subsection{Why the safe schedule differs: token commitment, not latent error}\label{sec:mechanism}
\begin{figure}[t]
\centering
\includegraphics[width=\linewidth]{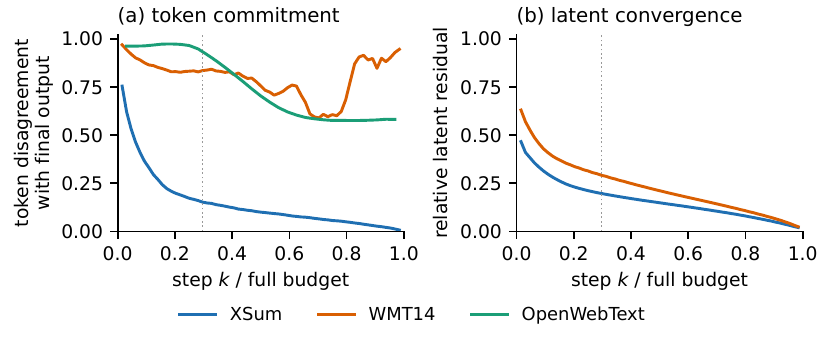}
\vspace{-14pt}
\caption{ELF-B trajectories (first 200 examples per task, canonical deterministic schedule). (a) Fraction of output positions whose intermediate decode differs from the final output. (b) Relative latent residual. Dotted: step 19.}
\label{fig:mechanism}
\vspace{-6pt}
\end{figure}

Error-optimized schedules aim to keep the latent trajectory close to the exact flow. \Cref{fig:mechanism} shows that this is not what separates the tasks. At step 19 the relative latent residual $\|\hat x_1(z_{t_k})-\hat x_1^{\rm final}\|/\|\hat x_1^{\rm final}\|$ is similar on XSum and WMT (0.20 vs 0.29), but 85\% of XSum output tokens already match the final output (half of all tokens never change after step 4 of 64), while WMT decodes stay truncated until the final step, even at step 63 where the latent is within 2.4\% of its endpoint (mean output length 2 tokens vs 18.5). On OWT, 58\% of output positions still differ from the final output at step 24 of 32. For XSum, later refinement even lowers ROUGE-L slightly, which is why \pol{prefix19} beats the full schedule. The matched-budget endpoint dependence index, $\mathrm{EDI}_K=\frac1n\sum_i[u_i(\pi^{\rm end}_K)-u_i(\pol{prefix}K)]$, summarizes this: for resampled minus prefix schedules it is negative at every $K$ on XSum ($-1.28$ at $K{=}8$ to $-0.14$ at $K{=}60$; \pol{H32}$-$\pol{prefix32}: $-0.21$ $[-0.34,-0.08]$, paired bootstrap) and positive at every $K$ on WMT ($+27$ to $+46$; \cref{app:edi}).

\begin{proposition}[Decoding margin]\label{prop:margin}
Let the decoder logits at a position, $\ell_v(x)$, be $L$-Lipschitz in the latent $x$ for every token $v$, and let $m(x)=\ell_{(1)}(x)-\ell_{(2)}(x)$ be the top-1/top-2 margin. If $\|x'-x\|<m(x)/(2L)$, the decoded token at $x'$ equals the one at $x$.
\end{proposition}
For a fixed decoding map, a prefix decode is therefore exact wherever the remaining drift is below the margin divided by $2L$, however large the absolute discretization error. The ELF decoder is time-conditioned, so on WMT the decode can also change because the decoding time changes; our measurements show that this, not residual latent error, dominates there.

\subsection{Distribution shift and monitoring}\label{sec:shift}
Applied unchanged, the certified ELF-B schedules stay within $\varepsilon=1$ on shifted data: XSum$\to$CNN/DailyMail \citep{hermann2015cnndm} \pol{prefix19} $\Delta=-0.17$ $[-0.36,+0.01]$ ROUGE-L; WMT14$\to$newstest2016 \pol{H32} $+0.09$ $[-0.03,+0.21]$ chrF; $\to$IWSLT17 \citep{cettolo2017iwslt} $+0.04$ $[-0.12,+0.22]$. Recalibrating on shifted data ($n=128$, $\varepsilon=1$) keeps \pol{prefix19} on CNN/DM in 379/500 splits (moving to the more conservative \pol{prefix26} at $\varepsilon=0.5$) and \pol{H32} on newstest2016 (491/500) and IWSLT17 (401/500), with 0/500 false certifications except 1/500 in one setting. As a recalibration trigger, a paired shadow monitor (an anytime-valid confidence sequence on $\Delta$ over shadowed requests; \citealp{waudbysmith2024estimating}) raised no false alarm on natural streams and flagged a synthetic $-3.5$ ROUGE-L degradation after a median of 79 requests with full shadowing (\cref{app:shift}).

\section{Related work}\label{sec:related}
\paragraph{Sampling theory and step selection.} Convergence theory bounds the error of diffusion samplers with approximate scores \citep{chen2023sampling,li2024towards,li2024accelerating,gatmiry2026high}, and higher-order methods reduce the number of steps \citep{li2025faster}. \citet{chen2024adaptive} choose time steps to minimize an error bound, Align Your Steps \citep{sabour2024align} optimizes schedules per model, and SharpEuler \citep{gupta2026sharpeuler} places flow-matching steps by velocity sharpness. These methods integrate to $t=1$ and control the terminal distribution; \tcda{} certifies decoded task utility, can stop before the endpoint, and treats such samplers as candidates (\cref{sec:baselines}). Collapse-and-refine \citep{huang2026collapse} separates the noise levels at which a denoiser refines the density on the data manifold from those at which it collapses onto it; our measurements suggest that strong conditioning (XSum) settles most of the decoded output early, after which the remaining steps change the latent far more than the tokens.
\paragraph{Discrete diffusion.} For masked diffusion, \citet{chen2026optimal} characterize optimal unmasking schedules information-theoretically, EB-Sampler bounds entropy per step \citep{benhamu2025ebsampler}, Fast-dLLM uses confidence thresholds and caching \citep{wu2025fastdllm}, Prophet commits answers early \citep{li2025prophet}, parallel-in-time sampling accelerates CTMC samplers \citep{yao2026parallel}, and DPRM orders tokens by a terminal-reward-induced process reward \citep{bu2026dprm}. They are per-example or theory-optimal samplers; \tcda{} decides, per deployment, which of them is safe at a given tolerance.
\paragraph{Adaptive computation and risk control.} Early exit adapts computation per input \citep{moon2024earlyexit,schuster2022calm}; learned stopping rules face train--serve skew, which we also observed (\cref{app:negative}). Risk-controlling calibration \citep{angelopoulos2021learn,bates2021rcps} and resampling-based multiple testing \citep{westfall1993,holm1979} supply the statistical tools; intersection--union tests \citep{berger1982iut} justify joint certification without extra multiplicity.

\section{Limitations}\label{sec:limits}
The guarantee is asymptotic and conditional on the library and on exchangeable calibration and deployment inputs; under shift, recalibration or monitoring is needed (\cref{sec:shift}). Guardrail contrasts with zero calibration variance are checked at the point estimate and are not covered by \cref{prop:fwer}. Guardrails measure cohort means and cannot catch rare per-example failures. For OWT we rely on automatic metrics (GPT-2 NLL, Gen.\ PPL, entropy, repetition); human judgments are not included. Certification costs $n\sum_\pi c(\pi)$ evaluations, which pays off after roughly 1,300--5,000 requests in our settings (\cref{app:economics}). Gains on OWT are modest ($1.33\times$), and for ELF-B OWT the deterministic grid degrades the full schedule, so we make no ELF-B OWT claim. All evidence uses automatic metrics on public benchmarks; tasks without a reliable utility metric need one before \tcda{} applies.

\section{Conclusion}
The cheapest quality-preserving schedule of a diffusion language model is a property of the task, the model and the sampler, not of the model alone: early decoding is best on XSum and catastrophic on WMT, early commit is free on WMT14 and ruinous on GSM8K, and the certified OWT topology changes with scale. \tcda{} turns this into a deployment decision with an error guarantee: it certifies utility and guardrails jointly over any library of samplers, falls back when evidence is lacking, and confirmed every prospective selection on fresh seeds. Token commitment, not latent error, explains which topology is safe.

\paragraph{Ethics statement.} We use public benchmarks and public model weights (T5, GPT-2, ELF, LLaDA) and one model we trained. No human-subject data were collected. Guardrails monitor relative output stability, not safety.
\paragraph{Reproducibility statement.} The supplement contains per-example results for every schedule, seed and task; the selector code with unit tests; the frozen protocols with their SHA-256 hashes; timing logs; and scripts that rebuild every table and figure.

\bibliographystyle{plainnat}
\bibliography{references}

\clearpage
\beginappendix
\crefalias{section}{appendix}

\section{Summary of changes from the submitted version}\label{app:changes}
\begin{itemize}
\item \textbf{Selector.} The submitted selector certified utility only, studentizing each bootstrap replicate by its own standard error; guardrails entered the reference policy but not the per-split certification. \tcda{} now certifies utility and guardrails jointly with fixed-denominator max-$T$ bounds (\cref{sec:tcda}, \cref{prop:fwer}); the submitted rule is kept as the ablation ``Rule U''. All reliability numbers come from one cohort of 500 calibration splits, reported as exact counts.
\item \textbf{Time grids.} ELF's reference sampler uses random time grids; all ELF-B results are recomputed with deterministic quantile grids and the prospective OWT test uses them (\cref{sec:grids}, \cref{app:grids}), including a repeat of the ELF-M XSum confirmation. XSum and WMT conclusions are unchanged. ELF-B OWT is no longer used for speed claims; OWT evidence comes from a prospective test on ELF-M and ELF-L.
\item \textbf{New evidence.} LLaDA-8B-Instruct on GSM8K, XSum and WMT14 (\cref{sec:llada}); prospective OWT at three scales (\cref{sec:prospective}); matched-cost baselines (\cref{sec:baselines}); a mechanically generated library (\cref{sec:elfb}); a mechanism analysis with a decoding-margin proposition (\cref{sec:mechanism}); distribution shift, recalibration and monitoring (\cref{sec:shift}).
\item \textbf{Definitions.} One reference policy $\pi^*(\varepsilon)$, defined with guardrails; the endpoint dependence index is defined at matched budget (\cref{app:edi}).
\item \textbf{Presentation.} Related work engages sampling theory and discrete-diffusion schedules; limitations are stated; sequential racing moves to \cref{app:racing}; internal audit logs are removed; hashes are in the supplementary README; figures use $\ge 9$\,pt text.
\end{itemize}

\section{Proofs}\label{app:proofs}
\paragraph{Setting.} Calibration examples are i.i.d.\ draws from the deployment distribution. For each candidate $\pi$ in a finite library $\Pi$, let $\mu_\pi=\E\Delta(\pi)$, $\sigma_\pi^2=\mathrm{Var}\,\Delta(\pi)$, and for each guardrail contrast $\eta_{\pi j}=\E h_j(\pi)$, $\sigma_{\pi j}^2=\mathrm{Var}\,h_j(\pi)$, all finite. Relative contrasts (e.g.\ length change divided by the full schedule's mean length) are smooth functions of means and are handled identically through their linearization (delta method).

\begin{proof}[Proof of \cref{prop:fwer}]
Let $\Pi_+=\{\pi:\sigma_\pi>0\}$ and $G_+=\{(\pi,j):\sigma_{\pi j}>0\}$. If $\sigma_\pi=0$, then $\hat s_\pi=0$ and $\pi$ is never certified. If $\sigma_{\pi j}=0$, the contrast is almost surely constant and its point check is exact. Candidates and contrasts with positive variance have $\hat s>0$ with probability tending to one. It therefore suffices to consider $\Pi_+$ and $G_+$.

\emph{Reduction.} Suppose the deployed $\hat\pi\ne\full$ lies outside $\Fe$. Then $\hat\pi$ was certified and either (a) $\mu_{\hat\pi}<-\varepsilon$, or (b) $\eta_{\hat\pi j}>\tau_j$ for some $j$. In case (a), $\hat\mu_{\hat\pi}-c_u\hat s_{\hat\pi}\ge-\varepsilon>\mu_{\hat\pi}$, so $M_u:=\max_{\pi\in\Pi_+}(\hat\mu_\pi-\mu_\pi)/\hat s_\pi>c_u$. In case (b), $\hat\eta_{\hat\pi j}+c_g\hat s_{\hat\pi j}\le\tau_j<\eta_{\hat\pi j}$, so $M_g:=\max_{(\pi,j)\in G_+}(\eta_{\pi j}-\hat\eta_{\pi j})/\hat s_{\pi j}>c_g$. Hence $\Pr(\hat\pi\notin\Fe)\le\Pr(M_u>c_u)+\Pr(M_g>c_g)$, whichever certified candidate is deployed.

\emph{Limits.} By the multivariate central limit theorem, $\sqrt n(\hat\mu-\mu)\Rightarrow Z\sim N(0,\Sigma)$, and $\sqrt n\,\hat s_\pi\to\sigma_\pi$ almost surely, so $M_u\Rightarrow M=\max_{\pi\in\Pi_+}Z_\pi/\sigma_\pi$ by continuous mapping. The nonparametric bootstrap of a vector of means is consistent \citep{bickel1981bootstrap}: conditionally on the data, $\sqrt n(\hat\mu^*-\hat\mu)\Rightarrow N(0,\Sigma)$ in probability, hence $M_u^*=\max_\pi(\hat\mu^*_\pi-\hat\mu_\pi)/\hat s_\pi\Rightarrow M$ and $c_u\to q_{1-\alpha_u}(M)$ in probability, $M$ being a maximum of non-degenerate Gaussians with a continuous distribution. Thus $\Pr(M_u>c_u)\to\alpha_u$. For the guardrails, $M_g\Rightarrow\max_{(\pi,j)}(-Z'_{\pi j}/\sigma_{\pi j})$, which has the same law as $\max_{(\pi,j)}Z'_{\pi j}/\sigma_{\pi j}$ because $Z'\overset{d}{=}-Z'$; the bootstrap quantity $\max(\hat\eta^*-\hat\eta)/\hat s$ converges to the same law, so $\Pr(M_g>c_g)\to\alpha_g$. Combining the two limits proves the claim.
\end{proof}
\paragraph{Remarks.} (i) In finite samples, the orientation of the guardrail bootstrap matters only through skewness; for right-skewed degeneration metrics (empty rate, repetition) the upper-tail quantile we use is the more conservative choice. (ii) In finite samples a contrast with positive variance can show zero calibration variance (e.g.\ no empty output in either schedule); it is then checked at its point estimate, which the guarantee does not cover. (iii) Our reliability study samples calibration sets without replacement from a finite pool and treats the pool as the population; the finite-population correction makes the bound slightly conservative there.

\begin{proof}[Proof of \cref{prop:margin}]
Let $v^\star=\arg\max_v\ell_v(x)$. For any $v\ne v^\star$, $\ell_{v^\star}(x')-\ell_v(x')\ge\ell_{v^\star}(x)-\ell_v(x)-2L\|x'-x\|\ge m(x)-2L\|x'-x\|>0$, so $v^\star$ remains the arg-max at $x'$.
\end{proof}

\section{Selector implementation and validation}\label{app:selector}
\paragraph{Algorithm.} Inputs: per-example utilities and guardrail metrics of every candidate and of $\full$ on $n$ calibration examples (shared noise), costs $c(\pi)$, tolerance $\varepsilon$, thresholds $\tau_j$, $\alpha_u=0.04$, $\alpha_g=0.01$, $B$. (1) Compute $\Delta_i(\pi)$, $\hat\mu_\pi$, $\hat s_\pi$; drop candidates with $\hat s_\pi=0$ from certification. (2) Form guardrail contrasts: a two-sided length guardrail is two one-sided contrasts; relative contrasts divide by the calibration mean of $\full$. (3) Draw $B$ index resamples (PCG64), compute $\max_\pi(\hat\mu^*_\pi-\hat\mu_\pi)/\hat s_\pi$ and $\max_{\pi,j}(\hat\eta^*_{\pi j}-\hat\eta_{\pi j})/\hat s_{\pi j}$ per resample, and take the $1-\alpha_u$ and $1-\alpha_g$ quantiles (\texttt{method='higher'}). (4) Certify, then deploy the cheapest certified candidate (ties: larger $\hat\mu_\pi$), else $\full$. We use $B=2{,}000$ in the 500-split studies and $B=10{,}000$ in the prospective test.
\paragraph{Ablations.} \emph{Rule U} (our re-implementation of the submitted selector): studentized max-$T$ on utility only, each replicate divided by its own standard error, zero-variance replicates set to $+\infty$, $\alpha=0.05$. \emph{Rule P}: Rule U plus guardrails checked at their calibration point estimates, which is what the submission described. \emph{Rule T}: a studentized joint rule over utility and guardrails, used as the primary rule of our first prospective run.
\paragraph{Validation.} Two independent implementations (a stand-alone reference and the project code) produce identical counts in all 108 ELF-B and 34 LLaDA settings at the same random seed, and agree within $\pm8/500$ under different seeds.

\section{Guardrails}\label{app:guardrails}
\begin{table}[h]
\centering\small
\caption{Guardrails. Length contrasts are relative to the full schedule's calibration mean; other contrasts are absolute differences of means.}
\label{tab:guardrails}
\begin{tabular}{lp{0.68\linewidth}}
\toprule
Task & Guardrail contrasts (candidate $-$ full schedule) and thresholds\\
\midrule
XSum (ELF, LLaDA) & length within $\pm10\%$; empty rate $\le+0.01$; source copy $\le+0.05$; novel trigrams $\ge-0.05$\\
WMT14 (ELF, LLaDA) & length ratio within $\pm0.10$; empty rate $\le+0.01$\\
OpenWebText & length within $\pm10\%$; distinct-2 $\ge-0.05$; adjacent repetition $\le+0.05$; decoder entropy $\le+0.5$; empty rate $\le+0.01$\\
GSM8K & answer-extraction failures $\le+0.01$\\
\bottomrule
\end{tabular}
\end{table}
\emph{Provenance.} The XSum and WMT thresholds were fixed before candidate selection. For ELF-B OWT in the submission, the repetition and entropy thresholds were introduced after degenerate loops were observed; the prospective OWT tests froze and hashed every threshold before generation.

\section{Complete reliability tables}\label{app:fulltables}
\begin{table}[h]
\centering\scriptsize\setlength{\tabcolsep}{2.6pt}
\caption{ELF-B, deterministic grids, core library: complete reliability grid (500 splits, $B=2{,}000$). Rule U is the submitted selector (studentized, utility only); Rule F is \tcda{}. Counts out of 500.}\label{tab:rel-core}
\begin{adjustbox}{max width=\linewidth}
\begin{tabular}{llllllrrllrr}
\toprule
 & & & & \multicolumn{4}{c}{Rule U} & \multicolumn{4}{c}{Rule F} \\ \cmidrule(lr){5-8} \cmidrule(lr){9-12}
Task & $n$ & $\varepsilon$ & $\pi^*(\varepsilon)$ & deployed (modal, count) & fallback & false & NFE & deployed (modal, count) & fallback & false & NFE \\
\midrule
XSum & 64 & 0 & \texttt{prefix19} & \texttt{canonical64} 450 & 450 & 2 & 1.07$\times$ & \texttt{canonical64} 452 & 452 & 0 & 1.07$\times$ \\
XSum & 64 & 0.25 & \texttt{prefix19} & \texttt{canonical64} 256 & 256 & 0 & 1.32$\times$ & \texttt{prefix19} 110 & 76 & 0 & 1.68$\times$ \\
XSum & 64 & 0.5 & \texttt{prefix19} & \texttt{canonical64} 156 & 156 & 0 & 1.62$\times$ & \texttt{prefix19} 229 & 5 & 0 & 2.42$\times$ \\
XSum & 64 & 1 & \texttt{prefix19} & \texttt{prefix19} 282 & 128 & 0 & 1.94$\times$ & \texttt{prefix19} 441 & 0 & 0 & 3.20$\times$ \\
XSum & 64 & 2 & \texttt{prefix19} & \texttt{prefix19} 354 & 111 & 0 & 2.14$\times$ & \texttt{prefix19} 500 & 0 & 0 & 3.37$\times$ \\
XSum & 128 & 0 & \texttt{prefix19} & \texttt{canonical64} 353 & 353 & 0 & 1.22$\times$ & \texttt{canonical64} 382 & 382 & 0 & 1.17$\times$ \\
XSum & 128 & 0.25 & \texttt{prefix19} & \texttt{prefix19} 261 & 44 & 0 & 2.25$\times$ & \texttt{prefix19} 257 & 18 & 0 & 2.36$\times$ \\
XSum & 128 & 0.5 & \texttt{prefix19} & \texttt{prefix19} 426 & 8 & 0 & 2.99$\times$ & \texttt{prefix19} 440 & 0 & 0 & 3.17$\times$ \\
XSum & 128 & 1 & \texttt{prefix19} & \texttt{prefix19} 485 & 3 & 0 & 3.26$\times$ & \texttt{prefix19} 499 & 0 & 0 & 3.37$\times$ \\
XSum & 128 & 2 & \texttt{prefix19} & \texttt{prefix19} 494 & 2 & 0 & 3.32$\times$ & \texttt{prefix19} 500 & 0 & 0 & 3.37$\times$ \\
XSum & 256 & 0 & \texttt{prefix19} & \texttt{prefix19} 204 & 202 & 0 & 1.63$\times$ & \texttt{canonical64} 229 & 229 & 0 & 1.53$\times$ \\
XSum & 256 & 0.25 & \texttt{prefix19} & \texttt{prefix19} 450 & 1 & 0 & 3.19$\times$ & \texttt{prefix19} 440 & 0 & 0 & 3.17$\times$ \\
XSum & 256 & 0.5 & \texttt{prefix19} & \texttt{prefix19} 499 & 0 & 0 & 3.36$\times$ & \texttt{prefix19} 500 & 0 & 0 & 3.37$\times$ \\
XSum & 256 & 1 & \texttt{prefix19} & \texttt{prefix19} 500 & 0 & 0 & 3.37$\times$ & \texttt{prefix19} 500 & 0 & 0 & 3.37$\times$ \\
XSum & 256 & 2 & \texttt{prefix19} & \texttt{prefix19} 500 & 0 & 0 & 3.37$\times$ & \texttt{prefix19} 500 & 0 & 0 & 3.37$\times$ \\
WMT & 64 & 0 & \texttt{H32} & \texttt{canonical64} 497 & 497 & 1 & 1.00$\times$ & \texttt{canonical64} 498 & 498 & 0 & 1.00$\times$ \\
WMT & 64 & 0.25 & \texttt{H32} & \texttt{canonical64} 232 & 232 & 6 & 1.25$\times$ & \texttt{H40} 135 & 124 & 5 & 1.40$\times$ \\
WMT & 64 & 0.5 & \texttt{H32} & \texttt{H32} 198 & 96 & 0 & 1.52$\times$ & \texttt{H32} 305 & 4 & 0 & 1.80$\times$ \\
WMT & 64 & 1 & \texttt{H32} & \texttt{H32} 334 & 27 & 0 & 1.81$\times$ & \texttt{H32} 429 & 0 & 0 & 1.99$\times$ \\
WMT & 64 & 2 & \texttt{H32} & \texttt{H32} 378 & 7 & 0 & 1.94$\times$ & \texttt{H32} 405 & 0 & 0 & 2.00$\times$ \\
WMT & 128 & 0 & \texttt{H32} & \texttt{canonical64} 493 & 493 & 0 & 1.01$\times$ & \texttt{canonical64} 495 & 495 & 0 & 1.00$\times$ \\
WMT & 128 & 0.25 & \texttt{H32} & \texttt{H32} 148 & 107 & 2 & 1.44$\times$ & \texttt{H32} 177 & 53 & 2 & 1.55$\times$ \\
WMT & 128 & 0.5 & \texttt{H32} & \texttt{H32} 352 & 11 & 0 & 1.82$\times$ & \texttt{H32} 428 & 0 & 0 & 1.94$\times$ \\
WMT & 128 & 1 & \texttt{H32} & \texttt{H32} 459 & 0 & 0 & 1.99$\times$ & \texttt{H32} 466 & 0 & 0 & 2.00$\times$ \\
WMT & 128 & 2 & \texttt{H32} & \texttt{H32} 464 & 0 & 0 & 2.00$\times$ & \texttt{H32} 464 & 0 & 0 & 2.00$\times$ \\
WMT & 256 & 0 & \texttt{H32} & \texttt{canonical64} 482 & 482 & 0 & 1.01$\times$ & \texttt{canonical64} 485 & 485 & 0 & 1.01$\times$ \\
WMT & 256 & 0.25 & \texttt{H32} & \texttt{H32} 307 & 9 & 1 & 1.78$\times$ & \texttt{H32} 334 & 3 & 0 & 1.82$\times$ \\
WMT & 256 & 0.5 & \texttt{H32} & \texttt{H32} 486 & 0 & 0 & 1.99$\times$ & \texttt{H32} 493 & 0 & 0 & 2.00$\times$ \\
WMT & 256 & 1 & \texttt{H32} & \texttt{H32} 485 & 0 & 0 & 2.00$\times$ & \texttt{H32} 485 & 0 & 0 & 2.00$\times$ \\
WMT & 256 & 2 & \texttt{H32} & \texttt{H32} 485 & 0 & 0 & 2.00$\times$ & \texttt{H32} 485 & 0 & 0 & 2.00$\times$ \\
OWT & 64 & 5\% & \texttt{uniform24} & \texttt{hybrid\_H16} 500 & 0 & 500 & 2.00$\times$ & \texttt{canonical32} 488 & 488 & 0 & 1.01$\times$ \\
OWT & 64 & 10\% & \texttt{uniform24} & \texttt{hybrid\_H16} 500 & 0 & 500 & 2.00$\times$ & \texttt{canonical32} 394 & 394 & 0 & 1.06$\times$ \\
OWT & 64 & 15\% & \texttt{uniform24} & \texttt{hybrid\_H16} 500 & 0 & 500 & 2.00$\times$ & \texttt{uniform24} 275 & 225 & 0 & 1.16$\times$ \\
OWT & 64 & 20\% & \texttt{uniform24} & \texttt{hybrid\_H16} 500 & 0 & 500 & 2.00$\times$ & \texttt{uniform24} 414 & 86 & 0 & 1.26$\times$ \\
OWT & 128 & 5\% & \texttt{uniform24} & \texttt{hybrid\_H16} 500 & 0 & 500 & 2.00$\times$ & \texttt{canonical32} 482 & 482 & 0 & 1.01$\times$ \\
OWT & 128 & 10\% & \texttt{uniform24} & \texttt{hybrid\_H16} 500 & 0 & 500 & 2.00$\times$ & \texttt{canonical32} 292 & 292 & 0 & 1.12$\times$ \\
OWT & 128 & 15\% & \texttt{uniform24} & \texttt{hybrid\_H16} 500 & 0 & 500 & 2.00$\times$ & \texttt{uniform24} 424 & 76 & 0 & 1.27$\times$ \\
OWT & 128 & 20\% & \texttt{uniform24} & \texttt{hybrid\_H16} 500 & 0 & 500 & 2.00$\times$ & \texttt{uniform24} 492 & 8 & 0 & 1.33$\times$ \\
OWT & 256 & 5\% & \texttt{uniform24} & \texttt{hybrid\_H16} 500 & 0 & 500 & 2.00$\times$ & \texttt{canonical32} 476 & 476 & 0 & 1.01$\times$ \\
OWT & 256 & 10\% & \texttt{uniform24} & \texttt{hybrid\_H16} 500 & 0 & 500 & 2.00$\times$ & \texttt{uniform24} 378 & 122 & 0 & 1.23$\times$ \\
OWT & 256 & 15\% & \texttt{uniform24} & \texttt{hybrid\_H16} 500 & 0 & 500 & 2.00$\times$ & \texttt{uniform24} 499 & 1 & 0 & 1.33$\times$ \\
OWT & 256 & 20\% & \texttt{uniform24} & \texttt{hybrid\_H16} 500 & 0 & 500 & 2.00$\times$ & \texttt{uniform24} 500 & 0 & 0 & 1.33$\times$ \\
\bottomrule
\end{tabular}
\end{adjustbox}
\end{table}

\begin{table}[h]
\centering\scriptsize\setlength{\tabcolsep}{2.6pt}
\caption{ELF-B, deterministic grids, dense library (127 candidates for XSum and WMT, 54 for OWT): complete reliability grid.}\label{tab:rel-dense}
\begin{adjustbox}{max width=\linewidth}
\begin{tabular}{llllllrrllrr}
\toprule
 & & & & \multicolumn{4}{c}{Rule U} & \multicolumn{4}{c}{Rule F} \\ \cmidrule(lr){5-8} \cmidrule(lr){9-12}
Task & $n$ & $\varepsilon$ & $\pi^*(\varepsilon)$ & deployed (modal, count) & fallback & false & NFE & deployed (modal, count) & fallback & false & NFE \\
\midrule
XSum & 64 & 0 & \texttt{prefix8} & \texttt{canonical64} 500 & 500 & 0 & 1.00$\times$ & \texttt{canonical64} 474 & 474 & 0 & 1.03$\times$ \\
XSum & 64 & 0.25 & \texttt{prefix7} & \texttt{canonical64} 499 & 499 & 0 & 1.00$\times$ & \texttt{hybrid\_H44\_bf75} 48 & 1 & 0 & 1.86$\times$ \\
XSum & 64 & 0.5 & \texttt{prefix6} & \texttt{canonical64} 499 & 499 & 0 & 1.00$\times$ & \texttt{hybrid\_H28\_bf75} 31 & 0 & 0 & 2.75$\times$ \\
XSum & 64 & 1 & \texttt{prefix5} & \texttt{canonical64} 499 & 499 & 0 & 1.00$\times$ & \texttt{prefix12} 58 & 0 & 0 & 4.77$\times$ \\
XSum & 64 & 2 & \texttt{prefix4} & \texttt{canonical64} 499 & 499 & 0 & 1.00$\times$ & \texttt{prefix8} 110 & 0 & 1 & 8.49$\times$ \\
XSum & 128 & 0 & \texttt{prefix8} & \texttt{canonical64} 489 & 489 & 0 & 1.01$\times$ & \texttt{canonical64} 411 & 411 & 0 & 1.12$\times$ \\
XSum & 128 & 0.25 & \texttt{prefix7} & \texttt{canonical64} 433 & 433 & 0 & 1.07$\times$ & \texttt{hybrid\_H24\_bf75} 33 & 0 & 0 & 2.62$\times$ \\
XSum & 128 & 0.5 & \texttt{prefix6} & \texttt{canonical64} 416 & 416 & 0 & 1.11$\times$ & \texttt{prefix9} 47 & 0 & 0 & 4.12$\times$ \\
XSum & 128 & 1 & \texttt{prefix5} & \texttt{canonical64} 406 & 406 & 0 & 1.15$\times$ & \texttt{prefix9} 130 & 0 & 0 & 6.75$\times$ \\
XSum & 128 & 2 & \texttt{prefix4} & \texttt{canonical64} 404 & 404 & 0 & 1.18$\times$ & \texttt{prefix6} 192 & 0 & 0 & 10.77$\times$ \\
XSum & 256 & 0 & \texttt{prefix8} & \texttt{canonical64} 325 & 325 & 0 & 1.27$\times$ & \texttt{canonical64} 259 & 259 & 0 & 1.41$\times$ \\
XSum & 256 & 0.25 & \texttt{prefix7} & \texttt{canonical64} 98 & 98 & 0 & 2.25$\times$ & \texttt{prefix16} 66 & 0 & 0 & 3.98$\times$ \\
XSum & 256 & 0.5 & \texttt{prefix6} & \texttt{canonical64} 93 & 93 & 0 & 2.77$\times$ & \texttt{prefix9} 111 & 0 & 0 & 5.81$\times$ \\
XSum & 256 & 1 & \texttt{prefix5} & \texttt{prefix7} 138 & 91 & 0 & 3.35$\times$ & \texttt{prefix7} 201 & 0 & 0 & 8.47$\times$ \\
XSum & 256 & 2 & \texttt{prefix4} & \texttt{prefix5} 223 & 91 & 0 & 3.89$\times$ & \texttt{prefix5} 324 & 0 & 0 & 12.60$\times$ \\
WMT & 64 & 0 & \texttt{hybrid\_H16\_bf75} & \texttt{canonical64} 499 & 499 & 0 & 1.00$\times$ & \texttt{canonical64} 496 & 496 & 2 & 1.01$\times$ \\
WMT & 64 & 0.25 & \texttt{hybrid\_H16\_bf75} & \texttt{canonical64} 335 & 335 & 0 & 1.09$\times$ & \texttt{canonical64} 41 & 41 & 1 & 1.59$\times$ \\
WMT & 64 & 0.5 & \texttt{hybrid\_H16\_bf75} & \texttt{canonical64} 277 & 277 & 0 & 1.18$\times$ & \texttt{hybrid\_H16\_bf75} 57 & 0 & 1 & 2.33$\times$ \\
WMT & 64 & 1 & \texttt{hybrid\_H16\_bf75} & \texttt{canonical64} 214 & 214 & 0 & 1.33$\times$ & \texttt{hybrid\_H16\_bf75} 195 & 0 & 3 & 3.43$\times$ \\
WMT & 64 & 2 & \texttt{resamp8} & \texttt{canonical64} 188 & 188 & 0 & 1.54$\times$ & \texttt{hybrid\_H16\_bf75} 267 & 0 & 2 & 4.05$\times$ \\
WMT & 128 & 0 & \texttt{hybrid\_H16\_bf75} & \texttt{canonical64} 496 & 496 & 0 & 1.00$\times$ & \texttt{canonical64} 494 & 494 & 0 & 1.01$\times$ \\
WMT & 128 & 0.25 & \texttt{hybrid\_H16\_bf75} & \texttt{canonical64} 146 & 146 & 0 & 1.31$\times$ & \texttt{hybrid\_H40\_bf25} 39 & 3 & 0 & 1.79$\times$ \\
WMT & 128 & 0.5 & \texttt{hybrid\_H16\_bf75} & \texttt{canonical64} 50 & 50 & 1 & 1.70$\times$ & \texttt{hybrid\_H16\_bf75} 112 & 0 & 0 & 2.77$\times$ \\
WMT & 128 & 1 & \texttt{hybrid\_H16\_bf75} & \texttt{hybrid\_H16\_bf75} 118 & 18 & 2 & 2.35$\times$ & \texttt{hybrid\_H16\_bf75} 275 & 0 & 1 & 3.81$\times$ \\
WMT & 128 & 2 & \texttt{resamp8} & \texttt{hybrid\_H16\_bf75} 201 & 6 & 0 & 3.22$\times$ & \texttt{hybrid\_H16\_bf75} 275 & 0 & 0 & 4.21$\times$ \\
WMT & 256 & 0 & \texttt{hybrid\_H16\_bf75} & \texttt{canonical64} 485 & 485 & 0 & 1.02$\times$ & \texttt{canonical64} 480 & 480 & 1 & 1.03$\times$ \\
WMT & 256 & 0.25 & \texttt{hybrid\_H16\_bf75} & \texttt{hybrid\_H16\_bf75} 61 & 16 & 0 & 1.88$\times$ & \texttt{hybrid\_H16\_bf75} 87 & 0 & 0 & 2.30$\times$ \\
WMT & 256 & 0.5 & \texttt{hybrid\_H16\_bf75} & \texttt{hybrid\_H16\_bf75} 189 & 3 & 0 & 2.81$\times$ & \texttt{hybrid\_H16\_bf75} 259 & 0 & 0 & 3.48$\times$ \\
WMT & 256 & 1 & \texttt{hybrid\_H16\_bf75} & \texttt{hybrid\_H16\_bf75} 348 & 0 & 0 & 3.73$\times$ & \texttt{hybrid\_H16\_bf75} 388 & 0 & 0 & 4.00$\times$ \\
WMT & 256 & 2 & \texttt{resamp8} & \texttt{hybrid\_H16\_bf75} 274 & 0 & 0 & 4.26$\times$ & \texttt{hybrid\_H16\_bf75} 229 & 0 & 0 & 4.51$\times$ \\
OWT & 64 & 5\% & \texttt{uniform24} & \texttt{prefix1} 500 & 0 & 500 & 32.00$\times$ & \texttt{uniform28} 246 & 246 & 0 & 1.07$\times$ \\
OWT & 64 & 10\% & \texttt{uniform24} & \texttt{prefix1} 500 & 0 & 500 & 32.00$\times$ & \texttt{uniform28} 350 & 87 & 0 & 1.14$\times$ \\
OWT & 64 & 15\% & \texttt{uniform20} & \texttt{prefix1} 500 & 0 & 500 & 32.00$\times$ & \texttt{uniform28} 278 & 20 & 0 & 1.21$\times$ \\
OWT & 64 & 20\% & \texttt{uniform20} & \texttt{prefix1} 500 & 0 & 500 & 32.00$\times$ & \texttt{uniform24} 346 & 3 & 0 & 1.29$\times$ \\
OWT & 128 & 5\% & \texttt{uniform24} & \texttt{prefix1} 500 & 0 & 500 & 32.00$\times$ & \texttt{uniform28} 408 & 87 & 0 & 1.12$\times$ \\
OWT & 128 & 10\% & \texttt{uniform24} & \texttt{prefix1} 500 & 0 & 500 & 32.00$\times$ & \texttt{uniform28} 349 & 11 & 0 & 1.19$\times$ \\
OWT & 128 & 15\% & \texttt{uniform20} & \texttt{prefix1} 500 & 0 & 500 & 32.00$\times$ & \texttt{uniform24} 391 & 0 & 0 & 1.29$\times$ \\
OWT & 128 & 20\% & \texttt{uniform20} & \texttt{prefix1} 500 & 0 & 500 & 32.00$\times$ & \texttt{uniform24} 421 & 0 & 0 & 1.35$\times$ \\
OWT & 256 & 5\% & \texttt{uniform24} & \texttt{prefix1} 500 & 0 & 500 & 32.00$\times$ & \texttt{uniform28} 492 & 1 & 0 & 1.15$\times$ \\
OWT & 256 & 10\% & \texttt{uniform24} & \texttt{prefix1} 500 & 0 & 500 & 32.00$\times$ & \texttt{uniform24} 315 & 0 & 0 & 1.26$\times$ \\
OWT & 256 & 15\% & \texttt{uniform20} & \texttt{prefix1} 500 & 0 & 500 & 32.00$\times$ & \texttt{uniform24} 497 & 0 & 0 & 1.33$\times$ \\
OWT & 256 & 20\% & \texttt{uniform20} & \texttt{prefix1} 500 & 0 & 500 & 32.00$\times$ & \texttt{uniform24} 362 & 0 & 0 & 1.40$\times$ \\
\bottomrule
\end{tabular}
\end{adjustbox}
\end{table}

\begin{table}[h]
\centering\scriptsize\setlength{\tabcolsep}{2.6pt}
\caption{ELF-B, original random-grid data of the submission (core library): complete reliability grid under the submitted rule (U) and \tcda{} (F).}\label{tab:rel-random}
\begin{adjustbox}{max width=\linewidth}
\begin{tabular}{llllllrrllrr}
\toprule
 & & & & \multicolumn{4}{c}{Rule U} & \multicolumn{4}{c}{Rule F} \\ \cmidrule(lr){5-8} \cmidrule(lr){9-12}
Task & $n$ & $\varepsilon$ & $\pi^*(\varepsilon)$ & deployed (modal, count) & fallback & false & NFE & deployed (modal, count) & fallback & false & NFE \\
\midrule
XSum & 64 & 0 & \texttt{prefix19} & \texttt{original64} 435 & 435 & 2 & 1.09$\times$ & \texttt{original64} 435 & 435 & 3 & 1.09$\times$ \\
XSum & 64 & 0.25 & \texttt{prefix19} & \texttt{original64} 249 & 249 & 0 & 1.36$\times$ & \texttt{prefix19} 121 & 114 & 0 & 1.65$\times$ \\
XSum & 64 & 0.5 & \texttt{prefix19} & \texttt{prefix19} 166 & 155 & 0 & 1.65$\times$ & \texttt{prefix19} 252 & 10 & 0 & 2.44$\times$ \\
XSum & 64 & 1 & \texttt{prefix19} & \texttt{prefix19} 296 & 110 & 0 & 2.00$\times$ & \texttt{prefix19} 468 & 0 & 0 & 3.27$\times$ \\
XSum & 128 & 0 & \texttt{prefix19} & \texttt{original64} 328 & 328 & 2 & 1.28$\times$ & \texttt{original64} 372 & 372 & 1 & 1.20$\times$ \\
XSum & 128 & 0.25 & \texttt{prefix19} & \texttt{prefix19} 283 & 58 & 0 & 2.24$\times$ & \texttt{prefix19} 266 & 34 & 0 & 2.33$\times$ \\
XSum & 128 & 0.5 & \texttt{prefix19} & \texttt{prefix19} 421 & 6 & 0 & 2.97$\times$ & \texttt{prefix19} 444 & 0 & 0 & 3.17$\times$ \\
XSum & 128 & 1 & \texttt{prefix19} & \texttt{prefix19} 485 & 0 & 0 & 3.28$\times$ & \texttt{prefix19} 500 & 0 & 0 & 3.37$\times$ \\
XSum & 256 & 0 & \texttt{prefix19} & \texttt{original64} 221 & 221 & 1 & 1.59$\times$ & \texttt{original64} 247 & 247 & 1 & 1.51$\times$ \\
XSum & 256 & 0.25 & \texttt{prefix19} & \texttt{prefix19} 455 & 0 & 0 & 3.21$\times$ & \texttt{prefix19} 456 & 0 & 0 & 3.22$\times$ \\
XSum & 256 & 0.5 & \texttt{prefix19} & \texttt{prefix19} 500 & 0 & 0 & 3.37$\times$ & \texttt{prefix19} 500 & 0 & 0 & 3.37$\times$ \\
XSum & 256 & 1 & \texttt{prefix19} & \texttt{prefix19} 500 & 0 & 0 & 3.37$\times$ & \texttt{prefix19} 500 & 0 & 0 & 3.37$\times$ \\
WMT & 64 & 0 & \texttt{H32} & \texttt{original64} 498 & 498 & 0 & 1.00$\times$ & \texttt{original64} 498 & 498 & 1 & 1.00$\times$ \\
WMT & 64 & 0.25 & \texttt{H32} & \texttt{original64} 297 & 297 & 5 & 1.19$\times$ & \texttt{original64} 217 & 217 & 4 & 1.29$\times$ \\
WMT & 64 & 0.5 & \texttt{H32} & \texttt{H32} 208 & 149 & 7 & 1.44$\times$ & \texttt{H32} 310 & 53 & 7 & 1.68$\times$ \\
WMT & 64 & 1 & \texttt{H32} & \texttt{H32} 359 & 51 & 0 & 1.74$\times$ & \texttt{H32} 454 & 2 & 0 & 1.97$\times$ \\
WMT & 128 & 0 & \texttt{H32} & \texttt{original64} 498 & 498 & 0 & 1.00$\times$ & \texttt{original64} 499 & 499 & 0 & 1.00$\times$ \\
WMT & 128 & 0.25 & \texttt{H32} & \texttt{H32} 179 & 176 & 0 & 1.37$\times$ & \texttt{H32} 209 & 135 & 0 & 1.44$\times$ \\
WMT & 128 & 0.5 & \texttt{H32} & \texttt{H32} 369 & 54 & 2 & 1.74$\times$ & \texttt{H32} 429 & 11 & 2 & 1.88$\times$ \\
WMT & 128 & 1 & \texttt{H32} & \texttt{H32} 472 & 9 & 0 & 1.95$\times$ & \texttt{H32} 486 & 0 & 0 & 2.00$\times$ \\
WMT & 256 & 0 & \texttt{H32} & \texttt{original64} 488 & 488 & 0 & 1.01$\times$ & \texttt{original64} 491 & 491 & 0 & 1.01$\times$ \\
WMT & 256 & 0.25 & \texttt{H32} & \texttt{H32} 274 & 101 & 0 & 1.56$\times$ & \texttt{H32} 301 & 62 & 0 & 1.64$\times$ \\
WMT & 256 & 0.5 & \texttt{H32} & \texttt{H32} 475 & 3 & 1 & 1.96$\times$ & \texttt{H32} 491 & 0 & 1 & 1.99$\times$ \\
WMT & 256 & 1 & \texttt{H32} & \texttt{H32} 497 & 0 & 0 & 2.00$\times$ & \texttt{H32} 496 & 0 & 0 & 2.00$\times$ \\
OWT & 64 & 0\% & \texttt{original32} & \texttt{original32} 500 & 500 & 0 & 1.00$\times$ & \texttt{original32} 500 & 500 & 0 & 1.00$\times$ \\
OWT & 64 & 5\% & \texttt{original32} & \texttt{original32} 500 & 500 & 0 & 1.00$\times$ & \texttt{original32} 500 & 500 & 0 & 1.00$\times$ \\
OWT & 64 & 10\% & \texttt{original32} & \texttt{original32} 500 & 500 & 0 & 1.00$\times$ & \texttt{original32} 500 & 500 & 0 & 1.00$\times$ \\
OWT & 64 & 20\% & \texttt{resamp\_K24} & \texttt{resamp\_K24} 271 & 210 & 19 & 1.18$\times$ & \texttt{resamp\_K24} 291 & 209 & 0 & 1.17$\times$ \\
OWT & 128 & 0\% & \texttt{original32} & \texttt{original32} 500 & 500 & 0 & 1.00$\times$ & \texttt{original32} 500 & 500 & 0 & 1.00$\times$ \\
OWT & 128 & 5\% & \texttt{original32} & \texttt{original32} 500 & 500 & 0 & 1.00$\times$ & \texttt{original32} 500 & 500 & 0 & 1.00$\times$ \\
OWT & 128 & 10\% & \texttt{original32} & \texttt{original32} 500 & 500 & 0 & 1.00$\times$ & \texttt{original32} 500 & 500 & 0 & 1.00$\times$ \\
OWT & 128 & 20\% & \texttt{resamp\_K24} & \texttt{resamp\_K24} 433 & 31 & 36 & 1.34$\times$ & \texttt{resamp\_K24} 463 & 37 & 0 & 1.30$\times$ \\
OWT & 256 & 0\% & \texttt{original32} & \texttt{original32} 500 & 500 & 0 & 1.00$\times$ & \texttt{original32} 500 & 500 & 0 & 1.00$\times$ \\
OWT & 256 & 5\% & \texttt{original32} & \texttt{original32} 500 & 500 & 0 & 1.00$\times$ & \texttt{original32} 500 & 500 & 0 & 1.00$\times$ \\
OWT & 256 & 10\% & \texttt{original32} & \texttt{original32} 500 & 500 & 0 & 1.00$\times$ & \texttt{original32} 500 & 500 & 0 & 1.00$\times$ \\
OWT & 256 & 20\% & \texttt{resamp\_K24} & \texttt{resamp\_K24} 445 & 0 & 55 & 1.38$\times$ & \texttt{resamp\_K24} 500 & 0 & 0 & 1.33$\times$ \\
\bottomrule
\end{tabular}
\end{adjustbox}
\end{table}

\section{LLaDA-8B-Instruct details}\label{app:llada}
\paragraph{Setup.} LLaDA-8B-Instruct with its default generation length and block settings per task; full schedules of 256 (GSM8K), 64 (XSum) and 128 steps (WMT14). The 21 candidates per task are early commit at several steps, uniform step reductions by 2, 4 and 8, front- and back-loaded allocations, a hybrid, Fast-dLLM ($\tau\in\{0.8,0.9,0.95\}$), EB-Sampler ($\gamma\in\{0.01,0.1,1\}$) and Prophet. At their published defaults, Fast-dLLM ($\tau=0.9$) uses $3.26\times$ fewer passes on GSM8K ($+0.6$ points), $6.6\times$ on WMT14 ($+0.03$ chrF) and $1.5\times$ on XSum ($-0.01$ ROUGE-L); EB-Sampler ($\gamma=0.1$) $1.9\times$, $4.1\times$ ($-1.05$ chrF) and $1.16\times$.
\paragraph{Prophet.} Our re-implementation of Prophet loses 26.3 points on GSM8K and 14.0 chrF on WMT14. To check it, we ran the official code (\cref{tab:prophet}): with its README settings it reproduces our numbers (51.5\% vs 50.5\% on 200 questions); with the evaluation script's answer-prefix constraint it reaches 73.0\% at $1.22\times$. In our setup both versions are dominated by Fast-dLLM and EB-Sampler; differences from the published results may stem from generation length and prompt format.
\begin{table}[h]
\centering\small
\caption{Prophet fidelity on GSM8K test questions 0--199 (LLaDA-8B-Instruct, 256 steps). \pol{prophet\_official\_unconstrained}: official code, README settings; \pol{prophet\_official\_constrained}: official evaluation script, which clamps answer positions to ``The answer is''.}\label{tab:prophet}
\begin{adjustbox}{max width=\linewidth}
\begin{tabular}{lrrr}
\toprule
Method & Accuracy (\%) & Mean NFE & NFE ratio \\
\midrule
\texttt{canonical} & 79.5 & 256.0 & 1.00$\times$ \\
\texttt{fd0.8} & 75.5 & 60.0 & 4.26$\times$ \\
\texttt{eb1.0} & 78.0 & 73.3 & 3.49$\times$ \\
\texttt{prophet\_simulated} & 50.5 & 190.4 & 1.34$\times$ \\
\texttt{prophet\_official\_unconstrained} & 51.5 & 185.3 & 1.38$\times$ \\
\texttt{prophet\_official\_constrained} & 73.0 & 209.6 & 1.22$\times$ \\
\bottomrule
\end{tabular}
\end{adjustbox}
\end{table}

\begin{table}[h]
\centering\scriptsize\setlength{\tabcolsep}{3pt}
\caption{LLaDA-8B-Instruct full-pool frontier (1{,}000 examples per task). \pol{fd}$\tau$: Fast-dLLM threshold; \pol{eb}$\gamma$: EB-Sampler; \pol{commitK}: unmask all after $K$ steps; \pol{uni}$k$, \pol{front}/\pol{back}: uniform, front- and back-loaded step allocations; \pol{prophet}: our re-implementation (\cref{tab:prophet}). Wall-clock: median of 7 interleaved repetitions on 200 examples.}\label{tab:app-llada-frontier}
\begin{adjustbox}{max width=\linewidth}
\begin{tabular}{llrrrlcr}
\toprule
Task & Schedule & NFE & NFE ratio & Utility & $\Delta$ [95\% CI] & Guardrails & Wall-clock \\
\midrule
GSM8K (EM) & \texttt{uni8} & 32.0 & 8.00$\times$ & 39.40 & -38.50 [-41.90, -34.90] & no &  \\
 & \texttt{commit32} & 32.0 & 8.00$\times$ & 10.20 & -67.70 [-70.70, -64.50] & yes & 7.68$\times$ \\
 & \texttt{fd0.8} & 60.4 & 4.24$\times$ & 77.20 & -0.70 [-2.50, +1.10] & yes & 4.25$\times$ \\
 & \texttt{back64} & 64.0 & 4.00$\times$ & 62.70 & -15.20 [-18.30, -12.00] & yes & 4.00$\times$ \\
 & \texttt{front64} & 64.0 & 4.00$\times$ & 58.50 & -19.40 [-22.80, -16.30] & no &  \\
 & \texttt{commit64} & 64.0 & 4.00$\times$ & 12.80 & -65.10 [-68.30, -61.80] & yes & 3.83$\times$ \\
 & \texttt{uni4} & 64.0 & 4.00$\times$ & 68.90 & -9.00 [-11.80, -6.10] & yes & 4.00$\times$ \\
 & \texttt{eb1.0} & 73.4 & 3.49$\times$ & 79.10 & +1.20 [-1.00, +3.40] & yes &  \\
 & \texttt{fd0.9} & 78.4 & 3.26$\times$ & 78.50 & +0.60 [-0.80, +2.00] & yes & 3.21$\times$ \\
 & \texttt{fd0.95} & 93.5 & 2.74$\times$ & 78.60 & +0.70 [-0.50, +2.00] & yes & 2.72$\times$ \\
 & \texttt{commit96} & 96.0 & 2.67$\times$ & 15.50 & -62.40 [-65.80, -59.20] & yes &  \\
 & \texttt{uni2} & 128.0 & 2.00$\times$ & 74.10 & -3.80 [-6.30, -1.40] & yes &  \\
 & \texttt{front128} & 128.0 & 2.00$\times$ & 72.90 & -5.00 [-7.80, -2.30] & yes &  \\
 & \texttt{back128} & 128.0 & 2.00$\times$ & 71.10 & -6.80 [-9.50, -4.10] & yes &  \\
 & \texttt{commit128} & 128.0 & 2.00$\times$ & 26.00 & -51.90 [-55.30, -48.60] & yes &  \\
 & \texttt{eb0.1} & 134.3 & 1.91$\times$ & 78.30 & +0.40 [-1.90, +2.50] & yes & 1.94$\times$ \\
 & \texttt{commit160} & 160.0 & 1.60$\times$ & 38.00 & -39.90 [-43.30, -36.70] & yes &  \\
 & \texttt{hyb} & 160.0 & 1.60$\times$ & 75.30 & -2.60 [-4.70, -0.60] & yes &  \\
 & \texttt{eb0.01} & 182.6 & 1.40$\times$ & 77.90 & +0.00 [-2.20, +2.20] & yes &  \\
 & \texttt{prophet} & 189.9 & 1.35$\times$ & 51.60 & -26.30 [-29.20, -23.30] & yes & 1.32$\times$ \\
 & \texttt{commit192} & 192.0 & 1.33$\times$ & 54.20 & -23.70 [-26.60, -20.80] & yes &  \\
 & \texttt{canonical} & 256.0 & 1.00$\times$ & 77.90 & +0.00 [+0.00, +0.00] & yes & 1.00$\times$ \\
\midrule
XSum (ROUGE-L) & \texttt{uni8} & 8.0 & 8.00$\times$ & 15.27 & -3.82 [-4.39, -3.24] & no &  \\
 & \texttt{commit8} & 8.0 & 8.00$\times$ & 16.78 & -2.31 [-2.81, -1.87] & no &  \\
 & \texttt{front16} & 16.0 & 4.00$\times$ & 14.78 & -4.30 [-4.87, -3.71] & no &  \\
 & \texttt{uni4} & 16.0 & 4.00$\times$ & 18.38 & -0.71 [-1.19, -0.24] & yes & 4.00$\times$ \\
 & \texttt{back16} & 16.0 & 4.00$\times$ & 18.87 & -0.21 [-0.60, +0.17] & yes & 4.00$\times$ \\
 & \texttt{commit16} & 16.0 & 4.00$\times$ & 18.60 & -0.49 [-0.88, -0.09] & no &  \\
 & \texttt{commit24} & 24.0 & 2.67$\times$ & 19.36 & +0.27 [-0.05, +0.58] & no &  \\
 & \texttt{commit32} & 32.0 & 2.00$\times$ & 19.26 & +0.17 [-0.06, +0.41] & yes & 1.85$\times$ \\
 & \texttt{uni2} & 32.0 & 2.00$\times$ & 19.13 & +0.05 [-0.32, +0.39] & yes & 2.00$\times$ \\
 & \texttt{front32} & 32.0 & 2.00$\times$ & 16.82 & -2.27 [-2.83, -1.75] & yes & 2.00$\times$ \\
 & \texttt{back32} & 32.0 & 2.00$\times$ & 19.03 & -0.06 [-0.38, +0.24] & yes & 2.00$\times$ \\
 & \texttt{fd0.8} & 37.2 & 1.72$\times$ & 19.01 & -0.07 [-0.14, -0.01] & yes & 1.70$\times$ \\
 & \texttt{commit40} & 40.0 & 1.60$\times$ & 18.75 & -0.34 [-0.52, -0.17] & yes &  \\
 & \texttt{hyb} & 40.0 & 1.60$\times$ & 19.05 & -0.03 [-0.19, +0.12] & yes &  \\
 & \texttt{eb1.0} & 42.4 & 1.51$\times$ & 18.90 & -0.19 [-0.48, +0.09] & yes &  \\
 & \texttt{fd0.9} & 42.8 & 1.50$\times$ & 19.08 & -0.01 [-0.05, +0.03] & yes & 1.47$\times$ \\
 & \texttt{fd0.95} & 46.8 & 1.37$\times$ & 19.08 & -0.01 [-0.05, +0.04] & yes & 1.35$\times$ \\
 & \texttt{commit48} & 48.0 & 1.33$\times$ & 18.93 & -0.16 [-0.29, -0.03] & yes &  \\
 & \texttt{prophet} & 53.5 & 1.20$\times$ & 18.94 & -0.15 [-0.31, -0.00] & yes & 1.09$\times$ \\
 & \texttt{eb0.1} & 55.0 & 1.16$\times$ & 18.90 & -0.19 [-0.48, +0.09] & yes & 1.17$\times$ \\
 & \texttt{eb0.01} & 60.1 & 1.06$\times$ & 18.90 & -0.19 [-0.48, +0.09] & yes &  \\
 & \texttt{canonical} & 64.0 & 1.00$\times$ & 19.09 & +0.00 [+0.00, +0.00] & yes & 1.00$\times$ \\
\midrule
WMT14 (chrF) & \texttt{prophet} & 7.0 & 18.31$\times$ & 41.86 & -14.04 [-14.81, -13.27] & yes & 16.81$\times$ \\
 & \texttt{fd0.8} & 15.6 & 8.23$\times$ & 55.69 & -0.21 [-0.41, -0.02] & yes & 7.53$\times$ \\
 & \texttt{uni8} & 16.0 & 8.00$\times$ & 50.98 & -4.92 [-5.56, -4.27] & yes & 8.00$\times$ \\
 & \texttt{commit16} & 16.0 & 8.00$\times$ & 46.52 & -9.37 [-10.02, -8.71] & yes & 7.82$\times$ \\
 & \texttt{fd0.9} & 19.4 & 6.58$\times$ & 55.93 & +0.03 [-0.08, +0.15] & yes & 6.03$\times$ \\
 & \texttt{eb1.0} & 19.6 & 6.53$\times$ & 54.99 & -0.91 [-1.28, -0.52] & yes &  \\
 & \texttt{fd0.95} & 22.6 & 5.68$\times$ & 55.95 & +0.05 [-0.00, +0.13] & yes & 5.23$\times$ \\
 & \texttt{eb0.1} & 31.0 & 4.13$\times$ & 54.85 & -1.05 [-1.47, -0.65] & yes & 3.88$\times$ \\
 & \texttt{uni4} & 32.0 & 4.00$\times$ & 54.47 & -1.42 [-1.84, -1.00] & yes &  \\
 & \texttt{commit32} & 32.0 & 4.00$\times$ & 55.33 & -0.57 [-0.70, -0.43] & yes &  \\
 & \texttt{front32} & 32.0 & 4.00$\times$ & 52.69 & -3.21 [-3.83, -2.64] & yes &  \\
 & \texttt{back32} & 32.0 & 4.00$\times$ & 52.81 & -3.08 [-3.61, -2.55] & yes &  \\
 & \texttt{eb0.01} & 36.9 & 3.47$\times$ & 54.85 & -1.05 [-1.47, -0.65] & yes &  \\
 & \texttt{commit48} & 48.0 & 2.67$\times$ & 55.56 & -0.34 [-0.43, -0.25] & yes &  \\
 & \texttt{commit64} & 64.0 & 2.00$\times$ & 55.89 & -0.00 [-0.01, +0.00] & yes &  \\
 & \texttt{front64} & 64.0 & 2.00$\times$ & 54.58 & -1.32 [-1.78, -0.91] & yes &  \\
 & \texttt{back64} & 64.0 & 2.00$\times$ & 54.59 & -1.31 [-1.69, -0.90] & yes &  \\
 & \texttt{uni2} & 64.0 & 2.00$\times$ & 55.62 & -0.27 [-0.61, +0.06] & yes &  \\
 & \texttt{hyb} & 80.0 & 1.60$\times$ & 55.90 & +0.00 [+0.00, +0.00] & yes & 1.60$\times$ \\
 & \texttt{commit80} & 80.0 & 1.60$\times$ & 55.89 & -0.00 [-0.01, +0.00] & yes &  \\
 & \texttt{commit96} & 96.0 & 1.33$\times$ & 55.90 & +0.00 [+0.00, +0.00] & yes &  \\
 & \texttt{canonical} & 128.0 & 1.00$\times$ & 55.90 & +0.00 [+0.00, +0.00] & yes & 1.00$\times$ \\
\bottomrule
\end{tabular}
\end{adjustbox}
\end{table}

\begin{table}[h]
\centering\scriptsize\setlength{\tabcolsep}{2.6pt}
\caption{LLaDA-8B-Instruct: complete reliability grid (500 splits, $B=2{,}000$). Rule P certifies utility only and checks guardrails at their point estimates; Rule F is \tcda{}.}\label{tab:app-llada-grid}
\begin{adjustbox}{max width=\linewidth}
\begin{tabular}{llllllrrllrr}
\toprule
 & & & & \multicolumn{4}{c}{Rule P} & \multicolumn{4}{c}{Rule F (\tcda{})} \\ \cmidrule(lr){5-8} \cmidrule(lr){9-12}
Task & $n$ & $\varepsilon$ & $\pi^*(\varepsilon)$ & deployed (modal) & fallback & false & NFE & deployed (modal) & fallback & false & NFE \\
\midrule
GSM8K & 128 & 1 pts & \texttt{fd0.8} & \texttt{canonical} 480 & 480 & 0 & 1.03$\times$ & \texttt{canonical} 477 & 477 & 0 & 1.03$\times$ \\
GSM8K & 128 & 2 pts & \texttt{fd0.8} & \texttt{canonical} 450 & 450 & 0 & 1.07$\times$ & \texttt{canonical} 410 & 410 & 1 & 1.14$\times$ \\
GSM8K & 128 & 5 pts & \texttt{fd0.8} & \texttt{canonical} 201 & 201 & 0 & 1.70$\times$ & \texttt{fd0.9} 162 & 83 & 0 & 2.36$\times$ \\
GSM8K & 256 & 1 pts & \texttt{fd0.8} & \texttt{canonical} 445 & 445 & 0 & 1.08$\times$ & \texttt{canonical} 453 & 453 & 0 & 1.07$\times$ \\
GSM8K & 256 & 2 pts & \texttt{fd0.8} & \texttt{canonical} 312 & 312 & 0 & 1.34$\times$ & \texttt{canonical} 319 & 319 & 0 & 1.32$\times$ \\
GSM8K & 256 & 5 pts & \texttt{fd0.8} & \texttt{fd0.8} 180 & 8 & 0 & 3.44$\times$ & \texttt{fd0.8} 166 & 5 & 0 & 3.47$\times$ \\
GSM8K & 512 & 1 pts & \texttt{fd0.8} & \texttt{canonical} 402 & 402 & 0 & 1.15$\times$ & \texttt{canonical} 414 & 414 & 0 & 1.13$\times$ \\
GSM8K & 512 & 2 pts & \texttt{fd0.8} & \texttt{fd0.9} 164 & 111 & 0 & 2.11$\times$ & \texttt{fd0.9} 154 & 132 & 0 & 1.99$\times$ \\
GSM8K & 512 & 5 pts & \texttt{fd0.8} & \texttt{fd0.8} 412 & 0 & 0 & 4.08$\times$ & \texttt{fd0.8} 398 & 0 & 0 & 4.05$\times$ \\
XSum & 128 & 0.5 & \texttt{back16} & \texttt{fd0.8} 320 & 12 & 4 & 1.69$\times$ & \texttt{fd0.8} 424 & 0 & 0 & 1.70$\times$ \\
XSum & 128 & 1 & \texttt{back16} & \texttt{fd0.8} 179 & 8 & 10 & 1.91$\times$ & \texttt{fd0.8} 341 & 0 & 0 & 1.80$\times$ \\
XSum & 256 & 0.5 & \texttt{back16} & \texttt{fd0.8} 256 & 0 & 1 & 1.84$\times$ & \texttt{fd0.8} 447 & 0 & 0 & 1.75$\times$ \\
XSum & 256 & 1 & \texttt{back16} & \texttt{commit32} 241 & 0 & 4 & 2.21$\times$ & \texttt{back32} 191 & 0 & 0 & 1.91$\times$ \\
WMT14 & 128 & 0.5 & \texttt{fd0.8} & \texttt{canonical} 421 & 421 & 0 & 1.15$\times$ & \texttt{fd0.9} 358 & 20 & 1 & 5.03$\times$ \\
WMT14 & 128 & 1 & \texttt{fd0.8} & \texttt{canonical} 395 & 395 & 0 & 1.22$\times$ & \texttt{fd0.8} 316 & 0 & 0 & 7.53$\times$ \\
WMT14 & 256 & 0.5 & \texttt{fd0.8} & \texttt{canonical} 463 & 463 & 0 & 1.07$\times$ & \texttt{fd0.9} 418 & 2 & 0 & 6.59$\times$ \\
WMT14 & 256 & 1 & \texttt{fd0.8} & \texttt{canonical} 459 & 459 & 0 & 1.08$\times$ & \texttt{fd0.8} 451 & 0 & 0 & 8.03$\times$ \\
\bottomrule
\end{tabular}
\end{adjustbox}
\end{table}

\section{Prospective tests}\label{app:e1}
\paragraph{Protocol.} Each OWT run froze a protocol file (library, utilities, guardrails and thresholds, tolerances of 5/10/15/20\% relative perplexity, rules, $B=10{,}000$ with a fixed PCG64 seed, seeds, success criteria) and recorded its SHA-256 before any generation; hashes are listed in the supplement. Stage 1 runs all candidates on 256 samples for a primary and a replicate noise seed; stage 2 runs every distinct non-fallback selection of the primary (\tcda{}) and secondary (Rule P) rules on $3\times1{,}000$ fresh samples. Success requires every per-seed one-sided 95\% bound and the hierarchical bound (resampling seeds, then samples within seed) to clear $-\varepsilon$, and every guardrail to pass at the point estimate and the \tcda{} upper bound. The library has 14 candidates per model (\pol{prefix}, \pol{hybrid}, \pol{resamp} and \pol{uniform} at 24--56 steps for ELF-M/L and 12--28 for ELF-B).
\paragraph{First run (random grids).} \Cref{tab:e1-random} lists all outcomes. Seven distinct schedules were selected; ELF-M \pol{prefix48}, ELF-L \pol{H48} (at 10--20\%) and ELF-B \pol{resamp28} (at 20\%) were confirmed, four were not. Because each generation seed fixed one random grid, stage 1 and stage 2 compared different grids; the grid study (\cref{app:grids}) quantifies this effect. The primary rule of that run was the studentized joint rule, whose degenerate critical values forced fallback for ELF-B.
\paragraph{Second run (deterministic grids).} \Cref{tab:e1det-stage1,tab:e1det-stage2} report all selections and outcomes. Every selection was confirmed.
\begin{table}[h]
\centering\scriptsize
\caption{First prospective OWT run (random time grids, seeds 300/301 for calibration, 400--402 for confirmation): every non-fallback selection and its outcome. T: studentized joint rule (primary); P: point guardrails (secondary).}\label{tab:e1-random}
\begin{adjustbox}{max width=\linewidth}
\begin{tabular}{lllll}
\toprule
Model & Policy & Tolerances & Rules & Outcome \\
\midrule
ELF-B-owt & \texttt{resamp\_K24} & 5\%, 10\%, 15\%, 20\% & P & failed \\
ELF-B-owt & \texttt{resamp\_K28} & 20\% & P & confirmed \\
ELF-L-owt & \texttt{hybrid\_H32} & 15\%, 20\% & P & failed \\
ELF-L-owt & \texttt{hybrid\_H48} & 5\% & P & failed \\
ELF-L-owt & \texttt{hybrid\_H48} & 10\%, 15\%, 20\% & T, P & confirmed \\
ELF-L-owt & \texttt{resamp\_K40} & 10\% & P & failed \\
ELF-M-owt & \texttt{prefix\_K48} & 5\%, 10\%, 15\%, 20\% & T, P & confirmed \\
ELF-M-owt & \texttt{resamp\_K40} & 15\%, 20\% & T, P & failed \\
\bottomrule
\end{tabular}
\end{adjustbox}
\end{table}

\begin{table}[h]
\centering\scriptsize
\caption{Prospective OWT, stage 1 (deterministic grids, seed 500, $B=10{,}000$): selections by rule. F: \tcda{} (primary); P: utility-only with point guardrails (secondary); U: utility only (comparison); T: the studentized joint rule of our first prospective run (comparison), whose critical values (77--306 for ELF-B and ELF-L, vs. 2.6 for F) force fallback.}\label{tab:e1det-stage1}
\begin{adjustbox}{max width=\linewidth}
\begin{tabular}{lrlllll}
\toprule
Model & $n$ & Tol. & F & P & U & T \\
\midrule
ELF-B-owt & 128 & 5\% & \texttt{uniform\_K28} & \texttt{uniform\_K28} & \texttt{resamp\_K12} & \texttt{canonical} \\
ELF-B-owt & 128 & 10\% & \texttt{uniform\_K24} & \texttt{uniform\_K24} & \texttt{resamp\_K12} & \texttt{canonical} \\
ELF-B-owt & 128 & 15\% & \texttt{uniform\_K24} & \texttt{uniform\_K24} & \texttt{resamp\_K12} & \texttt{canonical} \\
ELF-B-owt & 128 & 20\% & \texttt{uniform\_K24} & \texttt{uniform\_K24} & \texttt{resamp\_K12} & \texttt{canonical} \\
ELF-B-owt & 256 & 5\% & \texttt{uniform\_K28} & \texttt{uniform\_K28} & \texttt{resamp\_K12} & \texttt{canonical} \\
ELF-B-owt & 256 & 10\% & \texttt{uniform\_K24} & \texttt{uniform\_K24} & \texttt{resamp\_K12} & \texttt{canonical} \\
ELF-B-owt & 256 & 15\% & \texttt{uniform\_K24} & \texttt{uniform\_K24} & \texttt{resamp\_K12} & \texttt{canonical} \\
ELF-B-owt & 256 & 20\% & \texttt{uniform\_K24} & \texttt{uniform\_K24} & \texttt{resamp\_K12} & \texttt{canonical} \\
ELF-L-owt & 128 & 5\% & \texttt{hybrid\_H48} & \texttt{hybrid\_H48} & \texttt{hybrid\_H48} & \texttt{canonical} \\
ELF-L-owt & 128 & 10\% & \texttt{hybrid\_H48} & \texttt{hybrid\_H48} & \texttt{hybrid\_H48} & \texttt{canonical} \\
ELF-L-owt & 128 & 15\% & \texttt{hybrid\_H48} & \texttt{hybrid\_H48} & \texttt{hybrid\_H48} & \texttt{canonical} \\
ELF-L-owt & 128 & 20\% & \texttt{hybrid\_H48} & \texttt{hybrid\_H48} & \texttt{hybrid\_H48} & \texttt{canonical} \\
ELF-L-owt & 256 & 5\% & \texttt{hybrid\_H48} & \texttt{hybrid\_H48} & \texttt{hybrid\_H48} & \texttt{canonical} \\
ELF-L-owt & 256 & 10\% & \texttt{hybrid\_H48} & \texttt{hybrid\_H48} & \texttt{hybrid\_H48} & \texttt{canonical} \\
ELF-L-owt & 256 & 15\% & \texttt{hybrid\_H48} & \texttt{hybrid\_H48} & \texttt{hybrid\_H48} & \texttt{canonical} \\
ELF-L-owt & 256 & 20\% & \texttt{hybrid\_H48} & \texttt{hybrid\_H48} & \texttt{hybrid\_H48} & \texttt{canonical} \\
ELF-M-owt & 128 & 5\% & \texttt{prefix\_K48} & \texttt{prefix\_K48} & \texttt{prefix\_K48} & \texttt{prefix\_K48} \\
ELF-M-owt & 128 & 10\% & \texttt{prefix\_K48} & \texttt{prefix\_K48} & \texttt{prefix\_K48} & \texttt{prefix\_K48} \\
ELF-M-owt & 128 & 15\% & \texttt{prefix\_K48} & \texttt{prefix\_K48} & \texttt{prefix\_K48} & \texttt{prefix\_K48} \\
ELF-M-owt & 128 & 20\% & \texttt{prefix\_K48} & \texttt{prefix\_K48} & \texttt{prefix\_K48} & \texttt{prefix\_K48} \\
ELF-M-owt & 256 & 5\% & \texttt{prefix\_K48} & \texttt{prefix\_K48} & \texttt{prefix\_K48} & \texttt{canonical} \\
ELF-M-owt & 256 & 10\% & \texttt{prefix\_K48} & \texttt{prefix\_K48} & \texttt{prefix\_K48} & \texttt{canonical} \\
ELF-M-owt & 256 & 15\% & \texttt{prefix\_K48} & \texttt{prefix\_K48} & \texttt{prefix\_K48} & \texttt{prefix\_K48} \\
ELF-M-owt & 256 & 20\% & \texttt{prefix\_K48} & \texttt{prefix\_K48} & \texttt{prefix\_K48} & \texttt{prefix\_K48} \\
\bottomrule
\end{tabular}
\end{adjustbox}
\end{table}

\begin{table}[h]
\centering\scriptsize\setlength{\tabcolsep}{3pt}
\caption{Prospective OWT, stage 2: every distinct selection, seeds 600--602 ($3\times1{,}000$ samples). $\Delta$ in nats of $-$NLL; hierarchical one-sided 95\% bound; ANOVA $p$ for equal mean $\Delta$ across seeds; Gen.\ PPL of policy / full schedule. ELF-B rows are relative to a degraded deterministic full schedule (\cref{app:grids}) and are not speed claims.}\label{tab:e1det-stage2}
\begin{adjustbox}{max width=\linewidth}
\begin{tabular}{llrrrrrrlrl}
\toprule
Model & Policy & NFE & $\Delta_{600}$ & $\Delta_{601}$ & $\Delta_{602}$ & Hier.\ LCB & ANOVA $p$ & Gen.\ PPL & Wall & Outcome \\
\midrule
ELF-B-owt & \texttt{uniform\_K28} & 28 & +0.0789 & +0.0499 & +0.0488 & +0.0433 & 0.081 & 49.26 / 50.46 & 1.14$\times$ & confirmed \\
ELF-B-owt & \texttt{uniform\_K24} & 24 & -0.0098 & -0.0194 & -0.0334 & -0.0348 & 0.311 & 53.65 / 50.46 & 1.31$\times$ & confirmed \\
ELF-M-owt & \texttt{prefix\_K48} & 48 & -0.0055 & -0.0001 & -0.0007 & -0.0050 & 0.034 & 21.33 / 21.23 & 1.33$\times$ & confirmed \\
ELF-L-owt & \texttt{hybrid\_H48} & 48 & -0.0288 & -0.0224 & -0.0286 & -0.0302 & 0.121 & 23.68 / 23.03 & 1.33$\times$ & confirmed \\
\bottomrule
\end{tabular}
\end{adjustbox}
\end{table}

\paragraph{ELF-M XSum.}\label{app:elfm} We trained ELF-M on XSum from scratch (24 layers, hidden size 1{,}056, 342.5M parameters; Muon, global batch 512, 8$\times$B200, bf16, 39{,}800 steps, EMA 0.9999), froze the checkpoint that passed a predeclared quality gate, and preregistered a single candidate, \pol{prefix19}, against the 64-step schedule on untouched validation examples 1000--1999 with fresh seeds 126--128 at $\varepsilon=0.5$ ROUGE-L and four frozen guardrails. Results: per-seed $\Delta=+0.142, +0.191, +0.220$ (one-sided 95\% bounds $+0.003, +0.067, +0.101$); hierarchical $\Delta=+0.184$ (bound $+0.104$); all guardrails pass; 503.5\,s vs 154.5\,s per 1{,}000 examples ($3.26\times$). A one-way ANOVA across the three seeds finds no seed effect ($p=0.77$). \emph{Deterministic-grid repeat} (protocol frozen before generation; same checkpoint, candidate, $\varepsilon$, guardrails and criteria): on never-used IDs 2000--2999 with seeds 600--602, per-seed $\Delta=+0.390,+0.224,+0.341$, every one-sided bound $\ge+0.09$, hierarchical $\Delta=+0.319$ (bound $+0.217$), all guardrails pass; on the original IDs 1000--1999 and seeds 126--128, $\Delta=+0.139,+0.161,+0.229$, hierarchical $+0.176$ (bound $+0.094$), and per-seed $\Delta$ differs from the random-grid run by $-0.004,-0.031,+0.009$. Over 8 random grids with fixed noise (IDs 0--255) the SD of mean $\Delta$ is 0.042 against a within-run SE of 0.168 (ratio 0.25); the deterministic grid ($+0.133$) lies inside that spread. Per-step cost is unchanged (156.9\,s vs 506.6\,s per 1{,}000 examples).

\section{Time grids}\label{app:grids}
\begin{figure}[h]
\centering
\includegraphics[width=\linewidth]{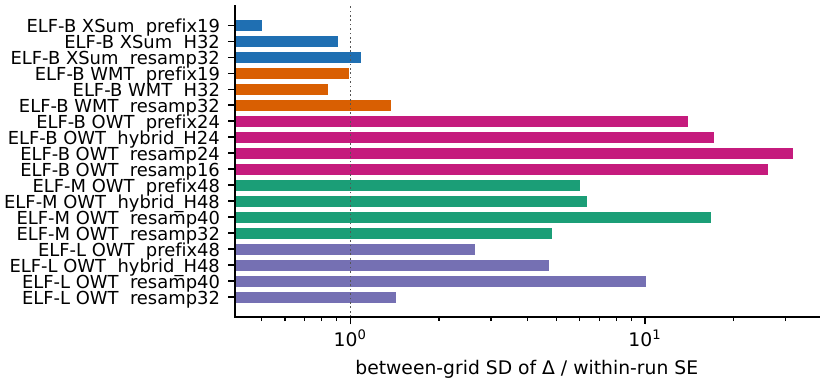}
\caption{Spread of policy effects across eight random time grids, in units of the within-run standard error (log scale). Values above 1 mean that the grid draw matters more than sampling noise.}
\label{fig:grids}
\end{figure}

\Cref{tab:grids} and \cref{fig:grids} report the grid-realization study. Holding the initial noise fixed (seed 124) and redrawing only the time grid changes policy effects by far more than their standard errors on OWT, while XSum and WMT are insensitive. \Cref{tab:canon-grid} compares the full schedule under the deterministic grid with eight random grids.
\begin{table}[h]
\centering\small
\caption{Full-schedule OWT quality, deterministic grid vs.\ 8 random grids (256 samples, noise seed 124). ELF-B random grid 1001 is degenerate.}
\label{tab:canon-grid}
\begin{tabular}{lll}
\toprule
Model & Deterministic grid: Gen.\ PPL (entropy) & Random grids: Gen.\ PPL range (entropy range)\\
\midrule
ELF-B & 51.3 (5.21) & 21.4--31.8 (4.87--5.19); grid 1001: 7.8 (2.88)\\
ELF-M & 21.5 (5.20) & 20.1--23.2 (5.16--5.21)\\
ELF-L & 23.3 (5.29) & 21.5--24.7 (5.24--5.31)\\
\bottomrule
\end{tabular}
\end{table}
\begin{table}[h]
\centering\scriptsize
\caption{Grid-realization study (noise seed 124 fixed, 256 examples, 8 random grids and the deterministic grid). $\Delta$: mean paired difference to the full schedule on the same grid; SD: across the 8 random grids; SE: mean within-run standard error. For full schedules the SD/SE column refers to the mean utility.}\label{tab:grids}
\begin{adjustbox}{max width=\linewidth}
\begin{tabular}{llrrrrrr}
\toprule
Model & Policy & NFE & $\Delta$ (det.) & mean $\Delta$ (random) & SD & SE & SD/SE \\
\midrule
ELF-B-xsum & \texttt{canonical64} & 64 & +0.000 & +0.000 & 0.127 & 0.866 & 0.15 \\
ELF-B-xsum & \texttt{prefix19} & 19 & +0.480 & +0.452 & 0.094 & 0.189 & 0.50 \\
ELF-B-xsum & \texttt{H32} & 32 & +0.227 & +0.108 & 0.088 & 0.097 & 0.91 \\
ELF-B-xsum & \texttt{resamp32} & 32 & -0.206 & -0.184 & 0.241 & 0.222 & 1.09 \\
ELF-B-de-en & \texttt{canonical64} & 64 & +0.000 & +0.000 & 0.178 & 0.989 & 0.18 \\
ELF-B-de-en & \texttt{prefix19} & 19 & -39.527 & -39.133 & 1.005 & 1.013 & 0.99 \\
ELF-B-de-en & \texttt{H32} & 32 & +0.182 & +0.052 & 0.107 & 0.127 & 0.84 \\
ELF-B-de-en & \texttt{resamp32} & 32 & -0.410 & -0.305 & 0.333 & 0.243 & 1.37 \\
ELF-B-owt & \texttt{canonical32} & 32 & +0.000 & +0.000 & 0.400 & 0.022 & 17.83 \\
ELF-B-owt & \texttt{prefix24} & 24 & -0.263 & -1.002 & 0.372 & 0.027 & 13.94 \\
ELF-B-owt & \texttt{hybrid\_H24} & 24 & +1.022 & +0.231 & 0.537 & 0.031 & 17.18 \\
ELF-B-owt & \texttt{resamp24} & 24 & +0.778 & -0.063 & 0.854 & 0.027 & 31.74 \\
ELF-B-owt & \texttt{resamp16} & 16 & +1.563 & +0.058 & 0.657 & 0.025 & 26.09 \\
ELF-M-owt & \texttt{canonical64} & 64 & +0.000 & +0.000 & 0.047 & 0.020 & 2.34 \\
ELF-M-owt & \texttt{prefix48} & 48 & +0.002 & -0.011 & 0.018 & 0.003 & 6.03 \\
ELF-M-owt & \texttt{hybrid\_H48} & 48 & -0.057 & -0.067 & 0.041 & 0.006 & 6.38 \\
ELF-M-owt & \texttt{resamp40} & 40 & -0.161 & -0.312 & 0.259 & 0.015 & 16.72 \\
ELF-M-owt & \texttt{resamp32} & 32 & -0.243 & -0.298 & 0.072 & 0.015 & 4.83 \\
ELF-L-owt & \texttt{canonical64} & 64 & +0.000 & +0.000 & 0.050 & 0.017 & 2.93 \\
ELF-L-owt & \texttt{prefix48} & 48 & -0.145 & -0.153 & 0.017 & 0.006 & 2.64 \\
ELF-L-owt & \texttt{hybrid\_H48} & 48 & -0.030 & -0.036 & 0.024 & 0.005 & 4.72 \\
ELF-L-owt & \texttt{resamp40} & 40 & -0.162 & -0.248 & 0.141 & 0.014 & 10.08 \\
ELF-L-owt & \texttt{resamp32} & 32 & -0.232 & -0.269 & 0.020 & 0.014 & 1.43 \\
\bottomrule
\end{tabular}
\end{adjustbox}
\end{table}

\section{Baselines}\label{app:baselines}
\paragraph{Error-optimized grid.} Following SharpEuler \citep{gupta2026sharpeuler}, we estimate the velocity sharpness $a(t)=\E\|v(z_{t_{k+1}})-v(z_{t_k})\|/(t_{k+1}-t_k)$ on 128 trajectories of the full deterministic schedule with a different noise seed (125), smooth it over 5 steps, floor it at 1\% of its maximum, set the step density $\rho(t)\propto a(t)^{1/2}$, and place $K+1$ grid points at the quantiles $i/K$ of its distribution function. The sampler is unchanged.
\paragraph{Heun.} Heun's method with $K$ evaluations ($K/2$ steps) on the logit-normal and on the uniform grid; self-conditioning uses the latest prediction.
\paragraph{Adaptive exit.} Simulated exactly from the recorded trajectories: \pol{stab}$s$ exits at the first step $\ge8$ at which the decoded tokens have not changed for $s$ steps; \pol{marg}$q$ exits when the 10th-percentile decoder top-1/top-2 margin exceeds the $q$-th percentile of that statistic at mid-trajectory on 128 profiling trajectories. Cost is the realized number of steps.
\begin{table}[h]
\centering\scriptsize
\caption{Matched-cost comparison on ELF-B (deterministic grids, full pools). \pol{sharpK}: error-optimized grid; \pol{heunK\_unif} / \pol{heunK}: Heun on the uniform / logit-normal grid; \pol{stab}$s$: exit after $s$ unchanged decodes; \pol{marg}$q$: exit when the 10th-percentile decoder margin exceeds the $q$-th percentile of its value at mid-trajectory on profiling data.}\label{tab:baselines}
\begin{adjustbox}{max width=\linewidth}
\begin{tabular}{llrrc}
\toprule
Task & Schedule & mean NFE & $\Delta$ vs full & Guardrails \\
\midrule
XSum & \texttt{prefix16} & 16.0 & +0.45 & yes \\
 & \texttt{prefix20} & 20.0 & +0.46 & yes \\
 & \texttt{prefix32} & 32.0 & +0.34 & yes \\
 & \texttt{prefix48} & 48.0 & +0.18 & yes \\
 & \texttt{H32} & 32.0 & +0.13 & yes \\
 & \texttt{H48} & 48.0 & +0.05 & yes \\
 & \texttt{resamp24} & 24.0 & -0.25 & yes \\
 & \texttt{resamp32} & 32.0 & -0.17 & yes \\
 & \texttt{uniform24} & 24.0 & -1.19 & yes \\
 & \texttt{uniform32} & 32.0 & -0.79 & yes \\
 & \texttt{sharp16} & 16.0 & -0.86 & yes \\
 & \texttt{sharp20} & 20.0 & -0.57 & yes \\
 & \texttt{sharp32} & 32.0 & -0.31 & yes \\
 & \texttt{sharp48} & 48.0 & -0.27 & yes \\
 & \texttt{heun16\_unif} & 16.0 & -2.13 & yes \\
 & \texttt{heun20\_unif} & 20.0 & -1.63 & yes \\
 & \texttt{heun32\_unif} & 32.0 & -0.91 & yes \\
 & \texttt{heun20} & 20.0 & -8.20 & yes \\
 & \texttt{heun32} & 32.0 & -7.01 & yes \\
 & \texttt{stab2} & 16.6 & +0.45 & yes \\
 & \texttt{stab4} & 23.5 & +0.33 & yes \\
 & \texttt{stab8} & 34.6 & +0.38 & yes \\
 & \texttt{marg50} & 33.9 & +0.26 & yes \\
 & \texttt{marg75} & 46.3 & +0.08 & yes \\
 & \texttt{marg90} & 53.9 & +0.07 & yes \\
\midrule
WMT14 & \texttt{hybrid\_H16\_bf75} & 16.0 & +0.22 & yes \\
 & \texttt{hybrid\_H20\_bf75} & 20.0 & +0.18 & yes \\
 & \texttt{H32} & 32.0 & +0.06 & yes \\
 & \texttt{H48} & 48.0 & +0.04 & yes \\
 & \texttt{resamp24} & 24.0 & -0.48 & yes \\
 & \texttt{resamp32} & 32.0 & -0.32 & yes \\
 & \texttt{uniform24} & 24.0 & -1.39 & yes \\
 & \texttt{uniform32} & 32.0 & -0.94 & yes \\
 & \texttt{sharp16} & 16.0 & -0.75 & yes \\
 & \texttt{sharp20} & 20.0 & -0.53 & yes \\
 & \texttt{sharp32} & 32.0 & -0.04 & yes \\
 & \texttt{sharp48} & 48.0 & +0.09 & yes \\
 & \texttt{heun16\_unif} & 16.0 & -2.33 & yes \\
 & \texttt{heun20\_unif} & 20.0 & -1.99 & yes \\
 & \texttt{heun32\_unif} & 32.0 & -1.12 & yes \\
 & \texttt{heun20} & 20.0 & -16.23 & yes \\
 & \texttt{heun32} & 32.0 & -12.46 & yes \\
 & \texttt{stab2} & 31.9 & -32.16 & no \\
 & \texttt{stab4} & 48.4 & -18.41 & no \\
 & \texttt{stab8} & 59.3 & -5.70 & no \\
 & \texttt{marg50} & 1.3 & -45.56 & no \\
 & \texttt{marg75} & 1.8 & -45.20 & no \\
 & \texttt{marg90} & 5.7 & -43.67 & no \\
\bottomrule
\end{tabular}
\end{adjustbox}
\end{table}

\begin{table}[h]
\centering\small
\caption{Wall-clock on 1{,}000 XSum examples (ELF-B, RTX PRO 6000, bf16; median and IQR of 7 interleaved repetitions).}\label{tab:timing}
\begin{adjustbox}{max width=\linewidth}
\begin{tabular}{lrrrrr}
\toprule
Schedule & NFE & NFE ratio & median s & IQR s & Wall-clock \\
\midrule
\texttt{canonical64} & 64.0 & 1.00$\times$ & 304.5 & 0.75 & 1.00$\times$ \\
\texttt{prefix19} & 19.0 & 3.37$\times$ & 93.0 & 0.18 & 3.28$\times$ \\
\texttt{H32} & 32.0 & 2.00$\times$ & 154.0 & 0.39 & 1.98$\times$ \\
\texttt{resamp32} & 32.0 & 2.00$\times$ & 154.0 & 0.23 & 1.98$\times$ \\
\texttt{sharp20} & 20.0 & 3.20$\times$ & 97.6 & 0.15 & 3.12$\times$ \\
\texttt{sharp32} & 32.0 & 2.00$\times$ & 154.0 & 0.31 & 1.98$\times$ \\
\texttt{heun20} & 20.0 & 3.20$\times$ & 97.8 & 0.27 & 3.11$\times$ \\
\texttt{heun32} & 32.0 & 2.00$\times$ & 154.2 & 0.35 & 1.98$\times$ \\
\texttt{stab4} & 23.5 & 2.72$\times$ & 161.9 & 0.44 & 1.88$\times$ \\
\texttt{marg75} & 46.2 & 1.39$\times$ & 357.8 & 0.06 & 0.85$\times$ \\
\bottomrule
\end{tabular}
\end{adjustbox}
\end{table}

\section{Mechanism and endpoint dependence}\label{app:edi}
\begin{figure}[h]
\centering
\includegraphics[width=\linewidth]{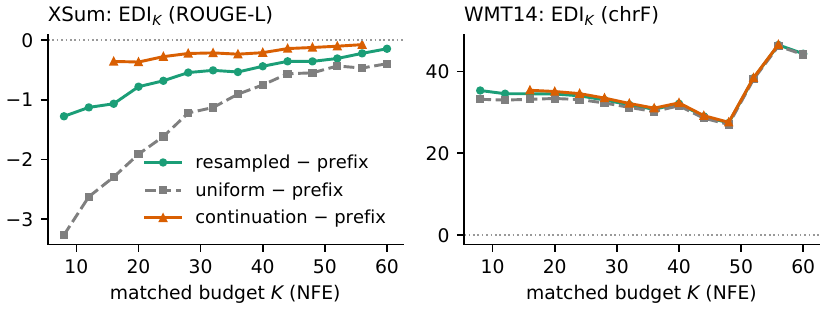}
\caption{Endpoint dependence at matched budget (ELF-B, deterministic grids, full pools).}
\label{fig:edi}
\end{figure}

For a budget $K$ and an endpoint-reaching schedule $\pi^{\rm end}_K$ (resampled, uniform or continuation), $\mathrm{EDI}_K=\frac1n\sum_i[u_i(\pi^{\rm end}_K)-u_i(\pol{prefix}K)]$ over paired examples with shared noise; intervals are 95\% paired percentile bootstrap intervals ($B=10{,}000$). Negative values mean that stopping early beats reaching the endpoint at the same cost. Matched-budget values (deterministic grids): XSum \pol{H32}$-$\pol{prefix32} $=-0.21$ $[-0.34,-0.08]$ ROUGE-L; WMT $+32.2$ $[31.6,32.8]$ chrF; OWT \pol{resamp24}$-$\pol{prefix24} $=+1.04$ $[1.00,1.08]$ nats ($+0.80$ $[0.78,0.82]$ on the original random-grid data). \Cref{fig:edi} shows the profile over $K$. On WMT, intermediate decodes are truncated at every step before the endpoint (mean length 2--13 tokens vs.\ 18.5 for the full schedule; up to 30\% empty), which makes the profile non-monotone above $K=48$.

\section{Distribution shift and monitoring}\label{app:shift}
\begin{table}[h]
\centering\small
\caption{Certified ELF-B schedules applied unchanged to shifted data (1{,}000 examples each).}\label{tab:shift}
\begin{adjustbox}{max width=\linewidth}
\begin{tabular}{llrrlc}
\toprule
Data & Schedule & Full & Schedule & $\Delta$ [95\% CI] & Guardrails \\
\midrule
xsum id & \texttt{prefix19} & 27.55 & 27.95 & +0.40 [+0.23, +0.57] & yes \\
cnndm shift & \texttt{prefix19} & 13.56 & 13.39 & -0.17 [-0.36, +0.01] & yes \\
wmt14 id & \texttt{H32} & 52.48 & 52.59 & +0.11 [-0.01, +0.25] & yes \\
wmt16 shift & \texttt{H32} & 58.19 & 58.28 & +0.09 [-0.03, +0.21] & yes \\
iwslt17 shift & \texttt{H32} & 52.80 & 52.84 & +0.04 [-0.12, +0.22] & yes \\
\bottomrule
\end{tabular}
\end{adjustbox}
\end{table}

\begin{table}[h]
\centering\small
\caption{Recalibration on shifted data with \tcda{} (core library, 500 splits).}\label{tab:recal}
\begin{adjustbox}{max width=\linewidth}
\begin{tabular}{lrrllrrr}
\toprule
Data & $n$ & $\varepsilon$ & $\pi^*(\varepsilon)$ & deployed (modal) & fallback & false & NFE \\
\midrule
xsum id & 128 & 0.5 & \texttt{prefix19} & \texttt{prefix19} 447 & 0 & 0 & 3.22$\times$ \\
xsum id & 128 & 1 & \texttt{prefix19} & \texttt{prefix19} 500 & 0 & 0 & 3.37$\times$ \\
xsum id & 256 & 0.5 & \texttt{prefix19} & \texttt{prefix19} 500 & 0 & 0 & 3.37$\times$ \\
xsum id & 256 & 1 & \texttt{prefix19} & \texttt{prefix19} 500 & 0 & 0 & 3.37$\times$ \\
cnndm shift & 128 & 0.5 & \texttt{prefix19} & \texttt{prefix26} 168 & 31 & 0 & 1.92$\times$ \\
cnndm shift & 128 & 1 & \texttt{prefix19} & \texttt{prefix19} 379 & 0 & 0 & 3.08$\times$ \\
cnndm shift & 256 & 0.5 & \texttt{prefix19} & \texttt{prefix26} 236 & 0 & 0 & 2.39$\times$ \\
cnndm shift & 256 & 1 & \texttt{prefix19} & \texttt{prefix19} 491 & 0 & 0 & 3.35$\times$ \\
wmt14 id & 128 & 0.5 & \texttt{H32} & \texttt{H32} 436 & 8 & 0 & 1.90$\times$ \\
wmt14 id & 128 & 1 & \texttt{H32} & \texttt{H32} 497 & 0 & 0 & 2.00$\times$ \\
wmt14 id & 256 & 0.5 & \texttt{H32} & \texttt{H32} 497 & 0 & 0 & 2.00$\times$ \\
wmt14 id & 256 & 1 & \texttt{H32} & \texttt{H32} 499 & 0 & 0 & 2.00$\times$ \\
wmt16 shift & 128 & 0.5 & \texttt{H32} & \texttt{H32} 415 & 7 & 1 & 1.90$\times$ \\
wmt16 shift & 128 & 1 & \texttt{H32} & \texttt{H32} 491 & 0 & 0 & 2.00$\times$ \\
wmt16 shift & 256 & 0.5 & \texttt{H32} & \texttt{H32} 494 & 0 & 0 & 1.99$\times$ \\
wmt16 shift & 256 & 1 & \texttt{H32} & \texttt{H32} 500 & 0 & 0 & 2.00$\times$ \\
iwslt17 shift & 128 & 0.5 & \texttt{H32} & \texttt{H32} 278 & 65 & 0 & 1.65$\times$ \\
iwslt17 shift & 128 & 1 & \texttt{H32} & \texttt{H32} 401 & 0 & 0 & 1.96$\times$ \\
iwslt17 shift & 256 & 0.5 & \texttt{H32} & \texttt{H32} 390 & 0 & 0 & 1.90$\times$ \\
iwslt17 shift & 256 & 1 & \texttt{H32} & \texttt{H32} 462 & 0 & 0 & 2.00$\times$ \\
\bottomrule
\end{tabular}
\end{adjustbox}
\end{table}

\paragraph{Monitoring.} The shadow monitor runs the full schedule on a random fraction $\rho$ of requests and maintains a predictable plug-in empirical-Bernstein lower confidence sequence \citep{waudbysmith2024estimating} on the rescaled paired difference over a sliding window of up to 300 shadowed requests, alarming when it falls below $-\varepsilon$. On natural streams (XSum$\to$CNN/DM, WMT14$\to$newstest2016, WMT14$\to$IWSLT17; 200 monitor seeds) it never alarmed. With a synthetic $-3.5$ ROUGE-L degradation injected after the shift, it alarmed in every run after a median of 79 requests at $\rho=1$, in 96\% of runs after a median of 62 shadowed requests at $\rho=0.1$, in 16\% at $\rho=0.05$ and never at $\rho=0.02$ within the stream length. A label-free two-sample KS test on source length flags XSum$\to$CNN/DM after 35 requests and IWSLT17 after 140, but it detects covariate shift, not quality loss.

\section{Calibration economics}\label{app:economics}
Calibrating costs $n\sum_{\pi}c(\pi)$ evaluations (including $\full$); deploying $\hat\pi$ saves $c(\full)-c(\hat\pi)$ per request. With $n=128$: XSum (ELF-B, core library) 59{,}008 evaluations, break-even after 1{,}312 requests with \pol{prefix19}; WMT 60{,}800 and 1{,}900 with \pol{H32}; ELF-M XSum (the 11-candidate library of the submission) 61{,}824 and 1{,}374; ELF-M/L OWT 79{,}872 and 4{,}992 with a 48-step schedule (counting every candidate separately; prefix and continuation candidates can reuse the full trajectory, which lowers the cost). When \tcda{} falls back, calibration is not recouped.

\section{Negative results}\label{app:negative}
Before \tcda{} we tried to learn when to stop. Learned anchor predictors (small heads and a LoRA adapter predicting the final latent) failed an offline cosine-similarity gate (best 0.840 vs.\ 0.85) and gave no generation gain with confidence intervals excluding zero; a task-label-free router found no reliable routing signal; and an ELF variant with heterogeneous per-token times trained on top of a frozen model did not support free-running commitment. These failures are consistent with train--serve skew and motivated a fixed library with certification.

\section{Sequential calibration}\label{app:racing}
A sequential variant evaluates nested calibration sizes $n_1<\dots<n_J$ ($J=6$), spends $\alpha_j=\alpha/J$ per stage (Bonferroni), maintains one-sided lower and upper simultaneous bounds, and stops when a candidate is certified or every candidate is shown infeasible. It is valid under optional stopping but conservative: median stopping sizes were 64--512 across settings with at most 0.4\% violations over 500 nested permutations. An earlier racing variant that stopped after one round, treated undecided candidates as unsafe and spent no $\alpha$ was invalid and is retired. We recommend fixed-$n$ calibration.

\end{document}